\pdfoutput=1

\documentclass[conference]{IEEEtran}

\usepackage{amsmath,epsfig,subfig}
\usepackage{url}
\usepackage{cite}
\usepackage{array}
\usepackage{hyperref}
\usepackage{xcolor}
\usepackage{adjustbox}
\usepackage[geometry]{ifsym}
\usepackage{enumitem}
\usepackage{comment}
\usepackage{multirow}
\usepackage{capt-of}
\usepackage{color,soul}
\usepackage{graphicx}

\newcommand\blfootnote[1]{%
  \begingroup
  \renewcommand\thefootnote{}\footnote{#1}%
  \addtocounter{footnote}{-1}%
  \endgroup
}

\usepackage[square,sort&compress,comma,numbers]{natbib}
\definecolor{blue_2}{RGB}{220, 240, 255}

\title{CURE-OR: Challenging Unreal and Real Environments for Object Recognition}

\author{\IEEEauthorblockN{Dogancan Temel*, Jinsol Lee* and Ghassan AlRegib}\\
\IEEEauthorblockA{\textit{Center for Signal and Information Processing} \\
\textit{Georgia Institute of Technology}, Atlanta, USA\\
\{cantemel, jinsol.lee, alregib\}@gatech.edu}}

\IEEEoverridecommandlockouts
\begin{document}\sloppy

\twocolumn[{%
\vspace{40mm}
{ \large
\begin{itemize}[leftmargin=2.5cm, align=parleft, labelsep=2.0cm, itemsep=4ex,]

\item[\textbf{Citation}]{D. Temel, J. Lee, and G. AlRegib, ``CURE-OR: Challenging unreal and real environments for object recognition,'' \textit{2018 17th IEEE International Conference on Machine Learning and Applications (ICMLA)}, Orlando, Florida, USA, 2018.}



\item[\textbf{Dataset}]{\url{https://ghassanalregib.com/cure-or/}}

\item[\textbf{Bib}]  {@INPROCEEDINGS\{Temel2018\_ICMLA,\\
author=\{D. Temel and J. Lee and G. AlRegib\},\\ 
booktitle=\{2018 17th IEEE International Conference on Machine Learning and Applications (ICMLA)\},\\
title=\{CURE-OR: Challenging unreal and real environments for object recognition\}, 
year={2018},\}
}

\item[\textbf{Copyright}]{\textcopyright 2018 IEEE. Personal use of this material is permitted. Permission from IEEE must be obtained for all other uses, in any current or future media, including reprinting/republishing this material for advertising or promotional purposes,
creating new collective works, for resale or redistribution to servers or lists, or reuse of any copyrighted component
of this work in other works. }

\item[\textbf{Contact}]{alregib@gatech.edu~~~~~~~\url{https://ghassanalregib.com/} \\dcantemel@gmail.com~~~~~~~\url{http://cantemel.com/}}
\end{itemize}
\thispagestyle{empty}
\newpage
\clearpage
\setcounter{page}{1}
}
}]

\twocolumn[{%
\renewcommand\twocolumn[1][]{#1}%
\maketitle
\vspace{-2mm}
\centering
\includegraphics[clip, trim=2mm 12.7cm 2mm 5mm, width=\linewidth]{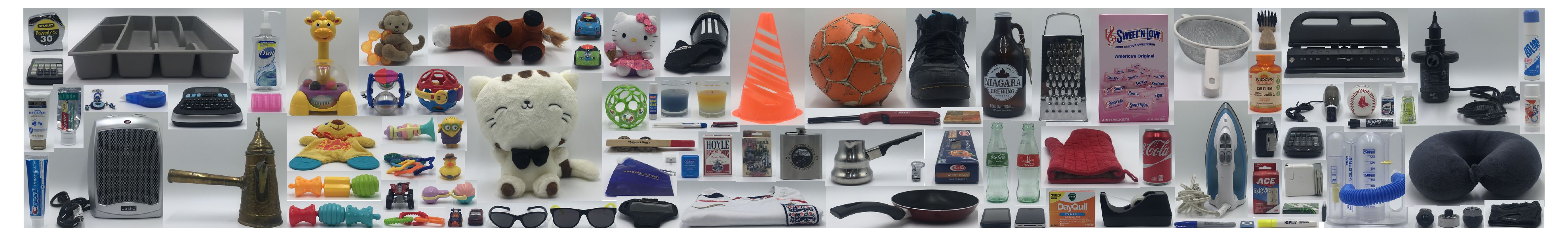}
\vspace{-3 mm}
\captionof{figure}{Objects in \texttt{CURE-OR} dataset that includes 1,000,000 images captured in controlled settings.}
\label{fig:objects}
\vspace{2mm}
}]

\def\x{{\mathbf x}}
\def\L{{\cal L}}


%
\begin{abstract}
In this paper, we introduce a large-scale, controlled, and multi-platform object recognition dataset denoted as Challenging Unreal and Real Environments for Object Recognition (\texttt{CURE-OR}). In this dataset, there are 1,000,000 images of 100 objects with varying size, color, and texture that are positioned in five different orientations and captured using five devices including a webcam, a DSLR, and three smartphone cameras in real-world (real) and studio (unreal) environments. The controlled challenging conditions include \emph{underexposure}, \emph{overexposure}, \emph{blur}, \emph{contrast}, \emph{dirty lens}, \emph{image noise}, \emph{resizing}, and \emph{loss of color information}. We utilize \texttt{CURE-OR} dataset to test recognition APIs---Amazon Rekognition and Microsoft Azure Computer Vision---and show that their performance significantly degrades under challenging conditions. Moreover, we investigate the relationship between object recognition and image quality and show that objective quality algorithms can estimate recognition performance under certain photometric challenging conditions. The dataset is publicly available at  \href{https://ghassanalregib.com/cure-or/}{https://ghassanalregib.com/cure-or/}. \blfootnote{*Equal contribution}   



\end{abstract}


\section{Introduction}
\label{sec:intro}
Recent advancements in algorithms and hardware platforms led to the development of state-of-the-art visual recognition methods that can achieve human-level performance in specific recognition tasks \cite{He2015,Wu2015}. Traditionally, these state-of-the-art recognition approaches were based on handcrafted features such as SIFT \cite{Lowe2004}, SURF \cite{Bay2006} and HoG \cite{Dalal2005}. However, recent data-driven methods started to outperform handcrafted methods as in the ImageNet Large Scale Visual Recognition Challenge (ILSVRC) \cite{Deng2009}. There are numerous datasets similar to ImageNet \cite{Li2004,Opelt2005,Everingham2005,Russell2008,Torralba2008,Krizhevsky09,Deng2009,Saenko2010,Lin2014}, whose images were obtained through photo sharing platforms and search engines. Even though these datasets can be comprehensive in terms of number of samples per categories, it is not possible to identify the relationship between specific conditions and recognition performance because of their uncontrolled nature. However, challenging conditions can significantly degrade the recognition performance as discussed in Section \ref{sec:exp}. Therefore, to investigate the robustness of recognition algorithms with respect to challenging conditions, we need to perform controlled experiments that include a comprehensive set of challenging condition types and levels.

Recent studies started investigating the effect of challenging conditions in images in terms of recognition performance \cite{Hosseini2017,Dodge2016,Zhou2017,Lu2017,Das2017,Temel2017_NIPSW}.  Hosseini \textit{et al.} \cite{Hosseini2017} evaluated the recognition performance of Google Cloud Vision API under Gaussian and impulse noise and showed that the API is vulnerable to tested challenging conditions. Dodge and Karam \cite{Dodge2016} evaluated the recognition performance of deep neural networks under blur, noise, contrast, and compression and showed that deep learning-based approaches are susceptible to degradation, particularly to blur and noise. Zhou \textit{et al.} \cite{Zhou2017} also studied the effect of blur and noise in deep network-based recognition and showed their vulnerabilities. Similarly, Lu \textit{et al.} \cite{Lu2017} and Das \textit{et al.} \cite{Das2017} showed the vulnerability of traffic sign recognition and detection systems in case of adversarial examples. Even though these studies shed light on a subset of conditions that can significantly affect algorithmic performance, the investigated challenging conditions are relatively limited. Specifically, aforementioned studies \cite{Hosseini2017,Dodge2016,Zhou2017,Lu2017,Das2017} focus on simulated challenging conditions and they overlook the possible imperfections in the acquisition process. Moreover, adversarial examples investigated in \cite{Lu2017,Das2017} are inherently different from realistic challenging scenarios because they are specifically tuned to bias the models. Temel \textit{et al.} \cite{Temel2017_NIPSW} focus on realistic challenging conditions but the application field is limited to traffic signs. Therefore, we need more comprehensive robustness analysis for generic object recognition tasks that considers realistic conditions as well as nonidealities in the acquisition processes.

\vspace{1mm}
This paper has three major components: the newly built \texttt{CURE-OR} dataset by which we eliminate the shortcomings of existing datasets; the performance benchmark of object recognition on the new dataset using the recognition APIs Amazon Rekognition and Microsoft Azure Computer Vision; and the investigation of the relationship between object recognition performance and full-reference image quality assessment, which measure the changes with respect to a challenge-free reference image. If we consider ideal conditions as reference conditions and non-ideal conditions as photometric challenging conditions, image quality should decrease under challenging conditions. \emph{Based on this assumption related to image quality, we hypothesize that object recognition performance under photometric challenging conditions can be estimated with objective image quality assessment algorithms.} Specifically, we focus on photometric challenging conditions including \emph{blurriness}, \emph{dirtiness of lens}, \emph{contrast}, \emph{overexposure}, \emph{undersexposure}, and \emph{sensor noise}. The contributions of this paper are five folds.

\vspace{2 mm}

\begin{itemize}[label=\textcolor{orange}{\FilledSmallSquare},leftmargin=*]   \setlength\itemsep{0.45em}
\item We introduce the most \emph{comprehensive} publicly-available object recognition dataset acquired with multiple device types under controlled challenging conditions in \emph{real-world} and \emph{photo studio} environments.

\item We provide a detailed description and a \emph{comparative} study of existing object datasets,  which  can  be  used as a source of reference in further studies.   

\item We provide a detailed \emph{analysis} of \emph{recognition APIs} in terms of recognition performance under challenging conditions.

\item We show that the performance of recognition applications significantly \emph{degrades} under challenging conditions.
    
\item We utilize full-reference objective \emph{image quality} assessment algorithms to \emph{estimate} the object recognition \emph{performance} under \emph{photometric} challenging conditions.
\end{itemize}

\vspace{2 mm}
 
\noindent{\bf Outline:}~We briefly discuss existing object datasets in Section~\ref{sec:related_work}. In Section~ \ref{sec:cure-or}, we introduce and describe the \texttt{CURE-OR} dataset. We analyze the performance of recognition APIs under challenging conditions in Section~\ref{sec:exp}, and we investigate the relationship between object recognition performance and objective quality assessment in Section~\ref{sec:quality}. Finally, we conclude our work in Section~\ref{sec:conc}.

\vspace{2mm}
\section{Object Datasets}
\label{sec:related_work}
\begin{table*}[]
\footnotesize

\centering
\caption{Main characteristics of publicly available object datasets and \texttt{CURE-OR}.}

\label{tab_datasets}
\begin{tabular}{>{\centering\arraybackslash}p{1.4cm}|>{\centering\arraybackslash}p{0.7cm}>{\centering\arraybackslash}p{1.1cm}>{\centering\arraybackslash}p{1.05cm}>{\centering\arraybackslash}p{1.4cm}>{\centering\arraybackslash}p{1.1cm}>{\centering\arraybackslash}p{1.0cm}>{\centering\arraybackslash}p{1.1cm}>{\centering\arraybackslash}p{1.4cm}>{\centering\arraybackslash}p{1.2cm}>{\centering\arraybackslash}p{1.3cm}}
\hline

 &   \textbf{\begin{tabular}[c]{@{}c@{}}Release \\ year\end{tabular}} & \textbf{\begin{tabular}[c]{@{}c@{}}Number\\ of\\ images\end{tabular}} & \textbf{Resolution} &  \textbf{\begin{tabular}[c]{@{}c@{}}Object \\ classes \end{tabular}} &\textbf{\begin{tabular}[c]{@{}c@{}}Number of\\ images per\\ class\end{tabular}}&\textbf{\begin{tabular}[c]{@{}c@{}}Number of\\different\\ objects per\\ class\end{tabular}} & \textbf{\begin{tabular}[c]{@{}c@{}}Number of \\ images per \\ same object\end{tabular}} & \textbf{\begin{tabular}[c]{@{}c@{}}Controlled \\ condition\\ (level)\end{tabular}} & \textbf{Background} & \textbf{Acquisition}  \\ \hline

\textbf{\href{http://www.cs.columbia.edu/CAVE/software/softlib/coil-100.php}{COIL100}}  & 1996   & 7,200 &128x128  &household  &\begin{tabular}[c]{@{}c@{}}all\\images \end{tabular}  & 100 &72  & \begin{tabular}[c]{@{}c@{}}angle (72)\\ \end{tabular}  &black  & \begin{tabular}[c]{@{}c@{}}CCD cam. \end{tabular}\\ \hline

\textbf{\href{http://www.cs.sfu.ca/~colour/image\_db/index.html}{SFU}}   & 1998   &\begin{tabular}[c]{@{}c@{}}all\\images \end{tabular} &637x468  &household &110 &11  &10  &\begin{tabular}[c]{@{}c@{}}lighting (5)\\ position (2) \end{tabular}  &black  & \begin{tabular}[c]{@{}c@{}}CCD cam.  \end{tabular} \\ \hline

\textbf{\href{http://www.cs.sfu.ca/~colour/image\_db/index.html}{SOIL-47}}   &2002   &\begin{tabular}[c]{@{}c@{}}all\\images \end{tabular}  & \begin{tabular}[c]{@{}c@{}}288x360 \\ and\\ 576x720\end{tabular}  & household &1,974 & 47 & 42 &\begin{tabular}[c]{@{}c@{}}angle (21)\\ lighting (2)\end{tabular}  & black &\begin{tabular}[c]{@{}c@{}}CCD cam. \end{tabular} \\ \hline

\textbf{\href{http://calvin.inf.ed.ac.uk/datasets/ethz-toys/}{ETHZ Toys}}  &2004  &63&720x576 &household  &\begin{tabular}[c]{@{}c@{}}all\\images \end{tabular} &9 &1-8 &angle (1-8) & \begin{tabular}[c]{@{}c@{}}black (model)\\ generic (test)\end{tabular}&- \\ \hline

\textbf{\href{http://www.cs.nyu.edu/~ylclab/data/norb-v1.0/}{NORB}} & 2004  &194,400  &640x480  & \begin{tabular}[c]{@{}c@{}}toys:  \\airp., trucks \\cars, human fig. \\four-leg. anim. \end{tabular}  &38,880  &10  & 3,888 &  \begin{tabular}[c]{@{}c@{}}angle (9)\\ azimuth (18)\\ lighting (6)\end{tabular} & \begin{tabular}[c]{@{}c@{}}white\\ textured\end{tabular} &\begin{tabular}[c]{@{}c@{}}2 CCD \\ cameras\end{tabular}  \\ \hline

\textbf{\href{http://www.vision.caltech.edu/Image\_Datasets/Caltech101/}{CalTech 101}}   &2004   &9,145  &300x200  &101 generic  &40-800& - & - & NA  &generic  & \begin{tabular}[c]{@{}c@{}}Online\\ search\end{tabular}  \\ \hline

\textbf{\href{http://host.robots.ox.ac.uk/pascal/VOC/voc2005/index.html}{PASCAL VOC}}  &2005   &2,232  & \begin{tabular}[c]{@{}c@{}}110x75 \\ to\\ 1,280x960\end{tabular} & \begin{tabular}[c]{@{}c@{}}motorcycles\\ bicycles\\ people\\ cars\end{tabular}  &168-547 &-  &-  & NA & generic & \begin{tabular}[c]{@{}c@{}}Online\\ search \\ETHZ\\UIUC  \end{tabular}  \\ \hline

\textbf{\href{http://www.emt.tugraz.at/~pinz/data/GRAZ\_02/}{GRAZ-02}}  & 2005 &1,476& \begin{tabular}[c]{@{}c@{}}480x640 \\ 640x480\end{tabular}&\begin{tabular}[c]{@{}c@{}}person, bike\\cars, counter\end{tabular} &311-420 &- &- & NA &generic&-  \\ \hline

\textbf{\href{http://aloi.science.uva.nl/}{ALOI}}  &2005  &110,250&768x576 &small obj. &\begin{tabular}[c]{@{}c@{}}all\\images \end{tabular} &1,000&$>$100 & \begin{tabular}[c]{@{}c@{}} angle (72)\\ light. angle (24) \\ light. color (12)\end{tabular}&black &\begin{tabular}[c]{@{}c@{}}3 CCD \\ cameras\end{tabular}  \\ \hline


\textbf{\href{http://labelme.csail.mit.edu/Release3.0/browserTools/php/dataset.php}{LabelME}} & 2008 &\begin{tabular}[c]{@{}c@{}}11,845 pic.\\ 18,524 fram.\end{tabular}&\begin{tabular}[c]{@{}c@{}}high\\resolution \end{tabular} &\begin{tabular}[c]{@{}c@{}}4,210\\ Wordnet\\categories \end{tabular}&\begin{tabular}[c]{@{}c@{}}up to\\27,719  \end{tabular} &- & - &NA&generic & \begin{tabular}[c]{@{}c@{}}Online\\ search\end{tabular} \\ \hline

\textbf{\href{http://groups.csail.mit.edu/vision/TinyImages/}{Tiny}}  &2008  &79,302,017&32x32 &\begin{tabular}[c]{@{}c@{}} 75,062  \\Wordnet \\nonabstract\\nouns \end{tabular} &\begin{tabular}[c]{@{}c@{}} up to \\3,000\end{tabular}&- &- &NA  &generic&\begin{tabular}[c]{@{}c@{}}Online\\search\end{tabular}  \\ \hline

\textbf{\href{https://www.cs.toronto.edu/~kriz/cifar.html}{CIFAR10 CIFAR100}}  &2009  &60,000&32x32 &\begin{tabular}[c]{@{}c@{}} 10 subclass of \\ vehic. and anim.,\\100 subclass of \\ 20 superclass  \end{tabular} & \begin{tabular}[c]{@{}c@{}} 6,000, \\600\end{tabular}  &- &- &NA&generic& \begin{tabular}[c]{@{}c@{}}Online\\search\end{tabular}  \\ \hline

\textbf{\href{http://image-net.org/}{ImageNet}}  &  2009 &3,200,000  & \begin{tabular}[c]{@{}c@{}}400x350 \\ (avg.) \end{tabular}  & \begin{tabular}[c]{@{}c@{}}5,247 \\Wordnet\\ synsets\end{tabular}  & \begin{tabular}[c]{@{}c@{}}600 \\ (avg.) \end{tabular}&-   & - &NA &generic  & \begin{tabular}[c]{@{}c@{}}Online\\ search\end{tabular}   \\ \hline

\textbf{\href{https://sites.google.com/site/bosstimuli/}{BOSS}} &  2010 &480  &2,000x2,000  &household  & \begin{tabular}[c]{@{}c@{}}all\\images \end{tabular}& 480 & 1 & NA &white  &-  \\ \hline

\textbf{\href{https://people.eecs.berkeley.edu/~jhoffman/domainadapt/\#datasets_code}{Office}}  &2010  &\begin{tabular}[c]{@{}c@{}}2,817 \\498\\795 \end{tabular}  & \begin{tabular}[c]{@{}c@{}}300x300 \\1,000x1,000 \\152x152 to \\752x752 \end{tabular}&\begin{tabular}[c]{@{}c@{}} 31 product  \\ categ.  \end{tabular}  &\begin{tabular}[c]{@{}c@{}}36-100 \\7-31  \\11-43 \end{tabular} &\begin{tabular}[c]{@{}c@{}}- \\4-6  \\4-6  \end{tabular}  &\begin{tabular}[c]{@{}c@{}}- \\ 1-9 \\1-14 \end{tabular}&\begin{tabular}[c]{@{}c@{}}NA\end{tabular} & \begin{tabular}[c]{@{}c@{}}studio light \\office env. \\office env.  \end{tabular}&\begin{tabular}[c]{@{}c@{}}Amazon.com\\DSLR cam.\\webcam \end{tabular}\\ \hline

\textbf{\href{http://rll.berkeley.edu/bigbird/}{BigBIRD}} & 2014  &150,000  &\begin{tabular}[c]{@{}c@{}}1,280x1,024 \\ 4,272x2,848\end{tabular}  &household &\begin{tabular}[c]{@{}c@{}}all\\images \end{tabular} &125  &1,200  & angle (120)  &white  & \begin{tabular}[c]{@{}c@{}}DSLR cam.\\ depth sensors\end{tabular}  \\ \hline

\textbf{\href{http://mscoco.org/}{MS-COCO}}  & 2014 &328,000& &\begin{tabular}[c]{@{}c@{}}91 classes \\incl. PACAL \\VOC 2011 \end{tabular} &\begin{tabular}[c]{@{}c@{}}3,600\\(avg)\\\end{tabular}  &- &-  &NA&generic &\begin{tabular}[c]{@{}c@{}}Online\\ search\end{tabular} \\ \hline

\textbf{iLab-20M}  &2016  &21,798,480&960x720 &\begin{tabular}[c]{@{}c@{}}toys \\ \end{tabular} &\begin{tabular}[c]{@{}c@{}}all\\images \end{tabular}  &704 &1,320  &\begin{tabular}[c]{@{}c@{}}angle (8) \\azimuth (11)\\focus (3)\\ lighting (5)\\backgr. ($>$14) \end{tabular}&\begin{tabular}[c]{@{}c@{}}textured \\backdr. (125), \\solid color\\backdr. (7) \end{tabular} &\begin{tabular}[c]{@{}c@{}}webcam (11) \\ \end{tabular} \\ \hline

\textbf{CURE-OR}   & 2018  &1,000,000 & \begin{tabular}[c]{@{}c@{}}648x968 \\756x1,008\\ 480x640\\460x816\\726x1,292 \end{tabular} & \begin{tabular}[c]{@{}c@{}} pers. belong.\\ office supply\\ household\\ toy \\sports/entert.\\health/pers.care \end{tabular}&\begin{tabular}[c]{@{}c@{}}100,000-\\270,000\end{tabular}  & 10-27  &10,000  & \begin{tabular}[c]{@{}c@{}} orientation (5)\\ backgr. (5)\\ challenge (79)\end{tabular} & \begin{tabular}[c]{@{}c@{}}white backdr.\\textured \\backdr (2),\\ living room \\ office\end{tabular} & \begin{tabular}[c]{@{}c@{}}DSLR cam. \\webcam\\phone(x3)\end{tabular} \\ \hline


\end{tabular}
\end{table*}

There are numerous datasets in the literature that can be used to benchmark the performance of object recognition algorithms. We can classify these datasets into two categories as \emph{controlled} datasets and \emph{uncontrolled} datasets. In \emph{uncontrolled} 
datasets \cite{Li2004,Opelt2005, Everingham2005,Russell2008,Torralba2008,Krizhevsky09,Deng2009,Saenko2010,Lin2014}, images are commonly acquired through search engines or photo sharing platforms. \emph{Uncontrolled} datasets are advantageous because it is easier to obtain large-scale datasets that include varying environmental conditions. The variety of images in these \emph{uncontrolled} datasets enables a large-scale performance assessment of algorithms. However, because of their \emph{uncontrolled} nature, it is not possible to understand the specific factors that affect recognition performance in these datasets.

\begin{figure}[b]
    \centering \includegraphics[width=0.61\linewidth, trim= 35mm 48mm 35mm 40mm]{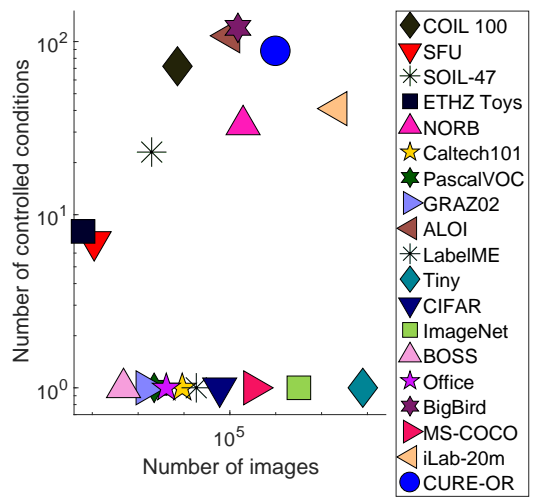} 
    \caption{An overview of object datasets in terms of dataset size and number of annotated controlled conditions.} 
    \label{fig:scatter_cure}
\end{figure}

In \emph{controlled} datasets \cite{Nene1996,Funt1998,Koubaroulis2002,Ferrari2004,LeCun2004,Geusebroek2005,Lai2011,Singh2014,Borji2016}, images are captured in a restricted setup under varying conditions. COIL-100 \cite{Nene1996}, SFU \cite{Funt1998}, SOIL-47 \cite{Koubaroulis2002}, and ETHZ Toy \cite{Ferrari2004} datasets contain up to thousands of images, which are limited compared to \emph{uncontrolled} recognition datasets. On the other hand, ALOI \cite{Geusebroek2005}, NORB \cite{LeCun2004}, RGB-D \cite{Lai2011}, and BigBIRD \cite{Singh2014} datasets are large-scale but their controlled conditions are limited to relative camera location, rotation, or lighting. Among all aforementioned \emph{controlled} datasets, iLab-20M \cite{Borji2016} is the largest dataset with a variety of acquisition conditions including relative camera location, focus level, scale, and background. In iLab-20M, one type of camera is used in the data acquisition, which makes the study overlook the domain differences that can be caused by acquisition device type as shown in \cite{Saenko2010}. Moreover, iLab-20M  is mainly used to test the invariance of learning-based approaches rather than analyzing factors that can affect the recognition performance, which is not possible to investigate until the data is publicly available. We summarize the main characteristics of existing datasets as well as introduced dataset \texttt{CURE-OR} in Table~\ref{tab_datasets} and compare these datasets in terms of number of images and annotated controlled conditions in Fig. \ref{fig:scatter_cure}. Because of the significant difference between the number of images in different datasets, we visualize data points in log scale. Ideally, a test dataset should be located on the top right corner so that it will be large-scale in terms of the number of images and annotated challenge types and levels. iLab-20M  and \texttt{CURE-OR}  are closest to top right with respect to compared datasets.





\section{CURE-OR Dataset}
\label{sec:cure-or}
\noindent{\bf Objects:}~\texttt{CURE-OR} dataset includes 100 objects of different color, size, and texture as shown in Fig. \ref{fig:objects}. The objects are grouped into 6 categories as toys, personal belongings, office supplies, household items, sports/entertainment items, and health/personal care items. The number of objects in each category varies between $10$ and $24$. For more details,  please refer to  \href{https://ghassanalregib.com/cure-or/}{https://ghassanalregib.com/cure-or/}.

\begin{figure}[htbp!]
    \begin{minipage}[t]{0.18\linewidth}
      \centering
    \includegraphics[width=\textwidth]{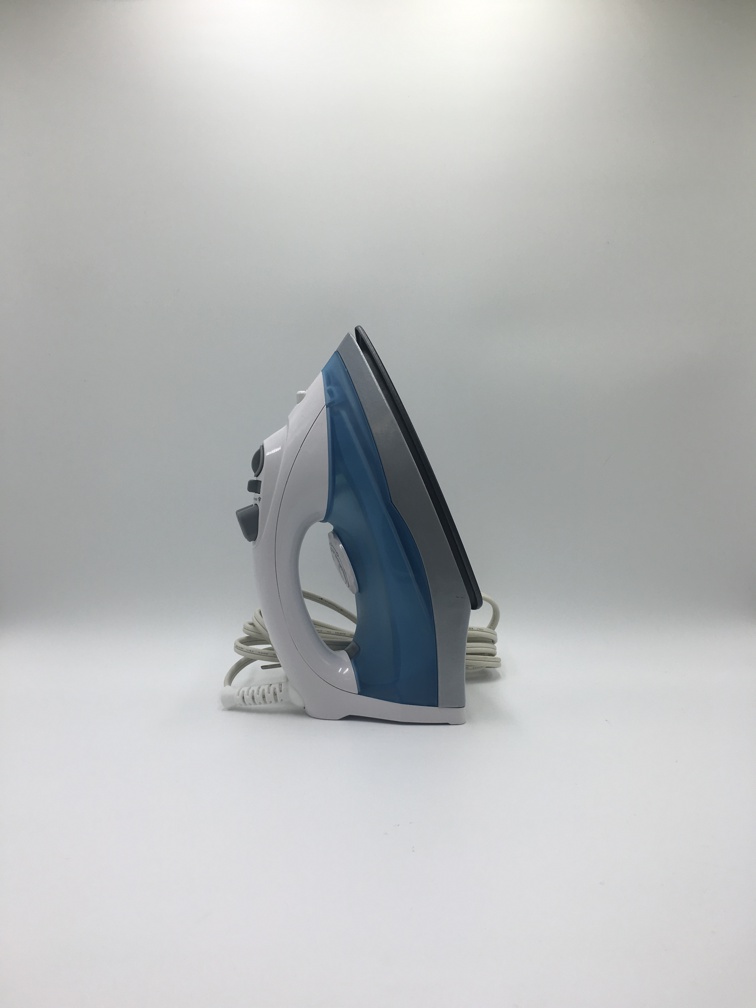}
      \vspace{0.01cm}
      \centerline{\footnotesize{(a)
    {\tabular[t]{@{}l@{}} White\\  \endtabular}}}
    \end{minipage}
    \hfill
    \begin{minipage}[t]{0.18\linewidth}
      \centering
    \includegraphics[width=\textwidth]{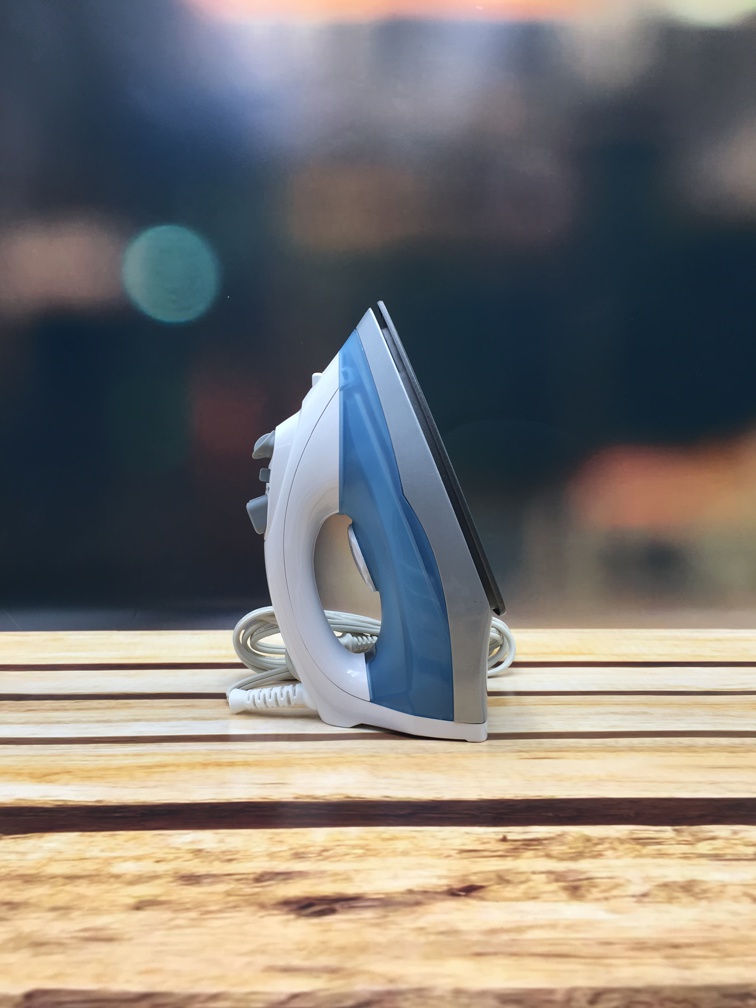}
      \vspace{0.01cm}
      \centerline{\footnotesize{(b)
      {\tabular[t]{@{}l@{}} Living\\ Room  \endtabular}}}
    \end{minipage}
    \hfill
    \begin{minipage}[t]{0.18\linewidth}
      \centering
    \includegraphics[width=\linewidth]{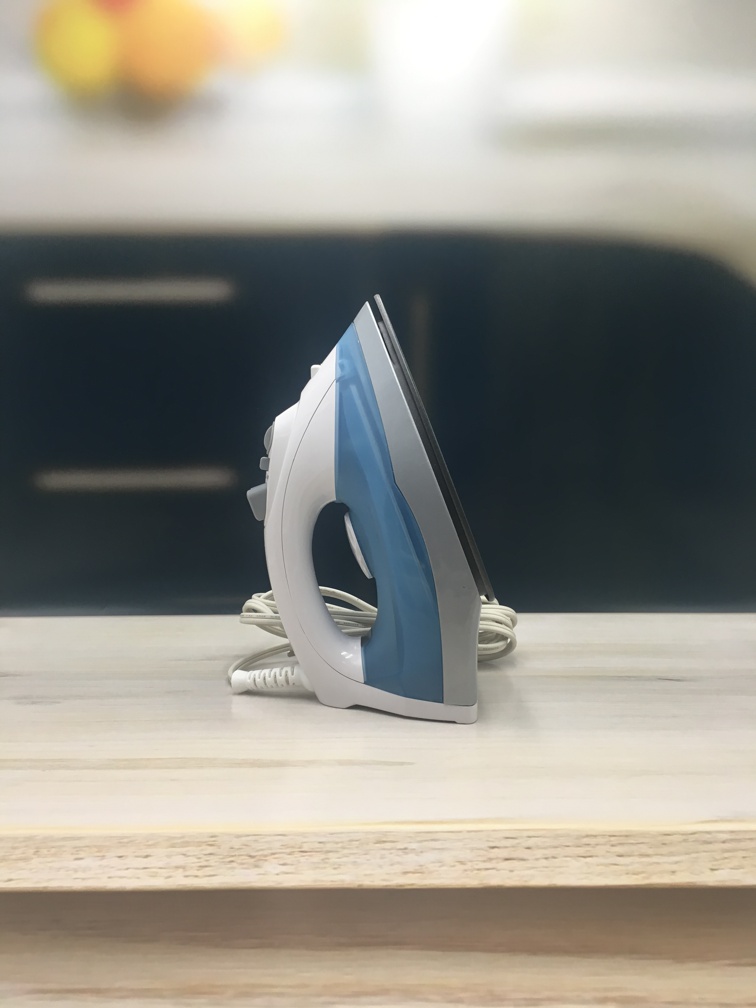}
      \vspace{0.01 cm}
      \centerline{\footnotesize{(c)
      {\tabular[t]{@{}l@{}} Kitchen\\ \endtabular}}}
    \end{minipage}
    \hfill
    \begin{minipage}[t]{0.18\linewidth}
      \centering
    \includegraphics[width=\linewidth]{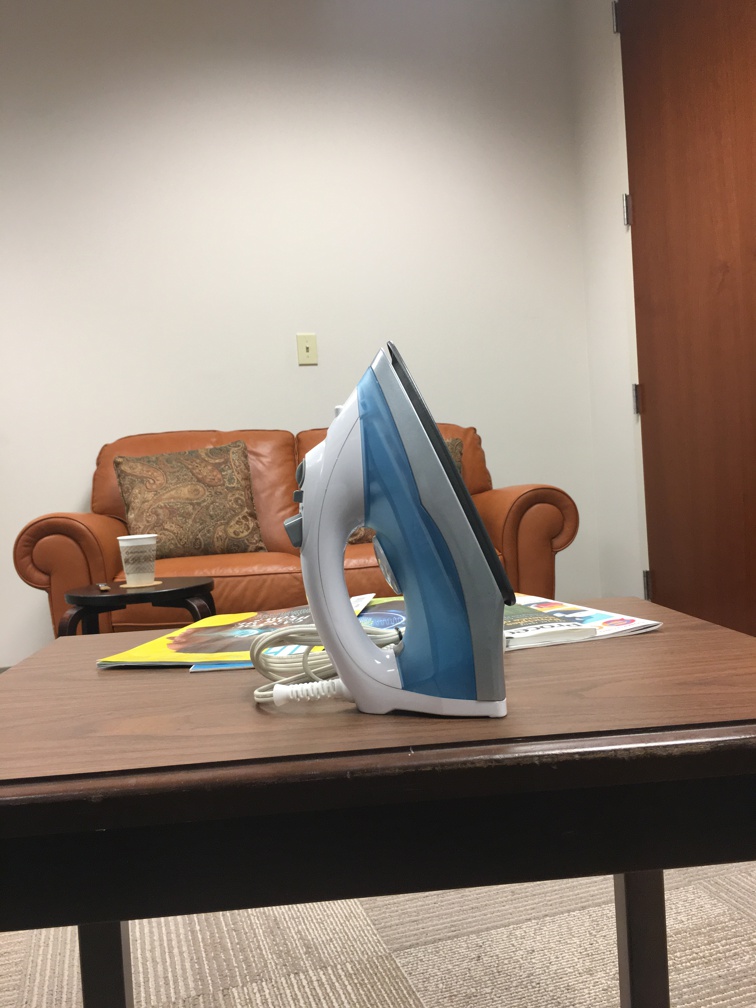}
      \vspace{0.01 cm}
      \centerline{\footnotesize{(d)
      {\tabular[t]{@{}l@{}} Living \\ Room  \endtabular}}}
    \end{minipage}
    \hfill
    \begin{minipage}[t]{0.18\linewidth}
      \centering
    \includegraphics[width=\linewidth]{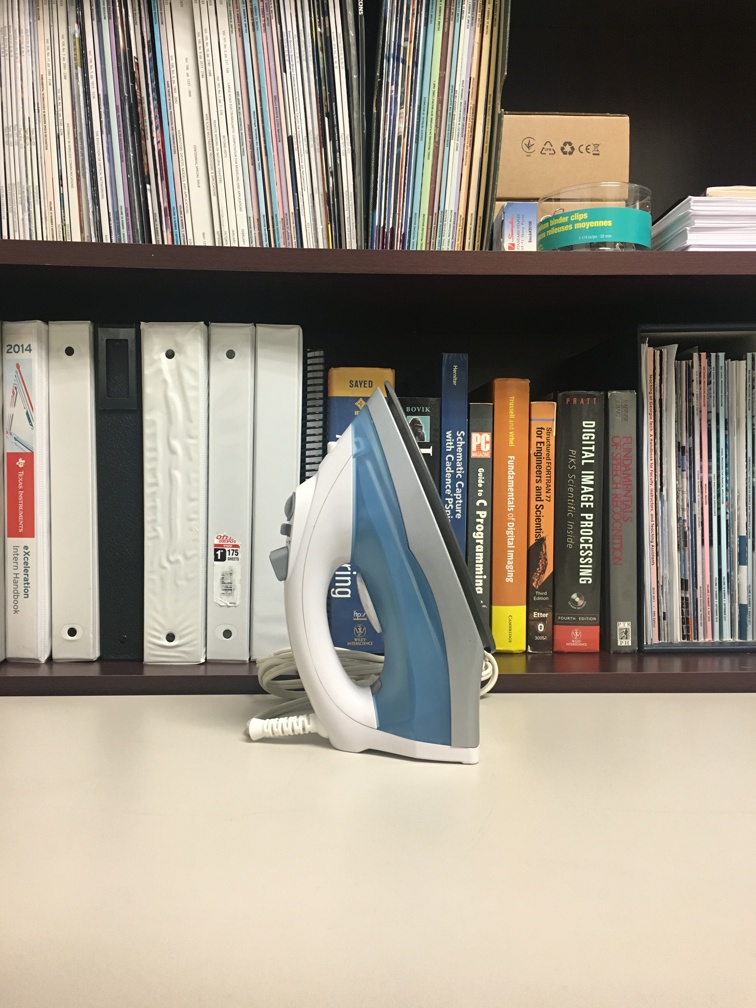}
      \vspace{0.01 cm}
      \centerline{\footnotesize{(e)
      {\tabular[t]{@{}l@{}} Office \\    \endtabular}}}
    \end{minipage}
    \centering
      \vspace{-1.0mm}
    \caption{ Studio (a)-(c) and real-world (d)-(e) environments.}
    \label{fig:backgrounds}
\end{figure}

\vspace{1mm}
\noindent{\bf Acquisition setup and image format:}~The images in \texttt{CURE-OR} were captured with 5 devices including iPhone 6s, HTC One X, LG Leon, Logitech C920 HD Pro Webcam, and Nikon D80, which were mounted on a Vanguard multi-mount horizontal bar attached to a tripod and controlled with wireless remotes. Each device had a slightly different perspective because of their relative locations. Images were captured in real and unreal environments. First, images were captured in a photo studio setup (LITEBOX Pro-240) with a white backdrop and two realistic backdrops of living room and kitchen. Second, images were captured in real-world environments including a living room and an office setup. The sample images of an object in every background are shown in Fig. \ref{fig:backgrounds}. Objects were captured at 5 different orientations: front, back, top, left-side, and right-side views. Originally, the width and height of captured images vary from $1,840$ to $5,168$ pixels with an average resolution of $2,457 \times 3,785 $. These images were scaled down to 1/4 of their original resolution and stored as JPEG to reduce dataset size.  

\vspace{-1mm}
\vspace{-4mm}

\newlength{\imgHeight}
\newlength{\imgWidth}
\newlength{\resWidth}
\newlength{\vblank}
\newlength{\cTypeHeight}
\newlength{\levelWidth}

\newcommand{\columnname}[1]
{\begin{minipage}[c][\cTypeHeight][t]{\imgWidth+\resWidth}\makebox[\imgWidth][c]{\textbf{\small#1}}\end{minipage}}

\newcommand{\rowname}[1]
{\begin{minipage}[b]{\levelWidth}
    \centering\rotatebox{90}{\makebox[\imgHeight][c]{\textbf{\small#1}}}
\end{minipage}}

\newcommand{\populateRow}[1]
{
    {\includegraphics[trim=0 0 0 3cm,clip,width=\imgWidth]{./Figs/api_example/2_5_2_069_03_#1.jpg}}\vspace{\vblank}\hfill
    {\includegraphics[trim=5mm 5mm 5mm 5mm,clip,height=\imgHeight]{./Figs/api_example_bar/03_#1.pdf}}\hfill
    {\includegraphics[trim=0 0 0 3cm,clip,width=\imgWidth]{./Figs/api_example/2_5_2_069_04_#1.jpg}}\hfill
    {\includegraphics[trim=5mm 5mm 5mm 5mm,clip,height=\imgHeight]{./Figs/api_example_bar/04_#1.pdf}}\hfill
    {\includegraphics[trim=0 0 0 9mm,clip,width=\imgWidth]{./Figs/api_example/2_5_2_069_05_#1.jpg}}\hfill
    {\includegraphics[trim=5mm 5mm 5mm 5mm,clip,height=\imgHeight]{./Figs/api_example_bar/05_#1.pdf}}\hfill
    {\includegraphics[trim=0 0 0 3cm,clip,width=\imgWidth]{./Figs/api_example/2_5_2_069_06_#1.jpg}}\hfill
    {\includegraphics[trim=5mm 5mm 5mm 5mm,clip,height=\imgHeight]{./Figs/api_example_bar/06_#1.pdf}}\hfill
    {\includegraphics[trim=0 0 0 9mm,clip,width=\imgWidth]{./Figs/api_example/2_5_2_069_07_#1.jpg}}\hfill
    {\includegraphics[trim=5mm 5mm 5mm 5mm,clip,height=\imgHeight]{./Figs/api_example_bar/07_#1.pdf}}\hfill
    {\includegraphics[trim=0 0 0 9mm,clip,width=\imgWidth]{./Figs/api_example/2_5_2_069_08_#1.jpg}}\hfill
    {\includegraphics[trim=5mm 5mm 5mm 5mm,clip,height=\imgHeight]{./Figs/api_example_bar/08_#1.pdf}}\hfill
    {\includegraphics[trim=0 0 0 3cm,clip,width=\imgWidth]{./Figs/api_example/2_5_2_069_09_#1.jpg}}\hfill
    {\includegraphics[trim=5mm 5mm 5mm 5mm,clip,height=\imgHeight]{./Figs/api_example_bar/09_#1.pdf}}\\
}

\begin{figure*}[!htbp]

    \setlength{\imgWidth}{.1\linewidth}
    \settoheight{\imgHeight}{\includegraphics[trim=0 0 0 3cm,clip,width=\imgWidth]
        {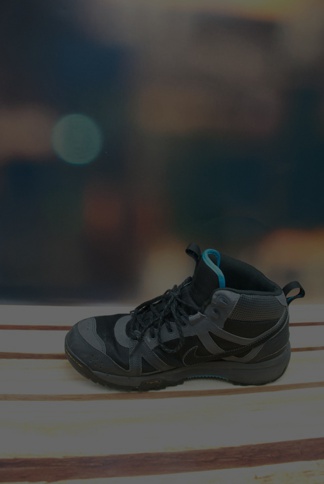}}
    \setlength{\resWidth}{.02\linewidth}
    \setlength{\vblank}{0.5mm}
    \setlength{\cTypeHeight}{4mm}
    \setlength{\levelWidth}{5mm}

    \begin{minipage}[b]{\levelWidth}\centering{\makebox[5mm][c]{}}\end{minipage}
        \columnname{Underexposure}\hfill
        \columnname{Overexposure}\hfill
        \columnname{Blur}\hfill
        \columnname{Contrast}\hfill
        \columnname{Dirty Lens 1}\hfill
        \columnname{Dirty Lens 2}\hfill
        \columnname{Salt \& Pepper}\\
    \rowname{Level 1}\populateRow{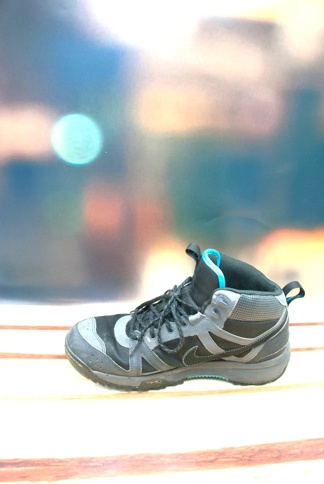}
    \rowname{Level 2}\populateRow{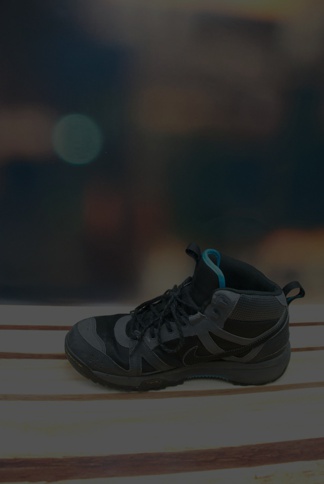}
    \rowname{Level 3}\populateRow{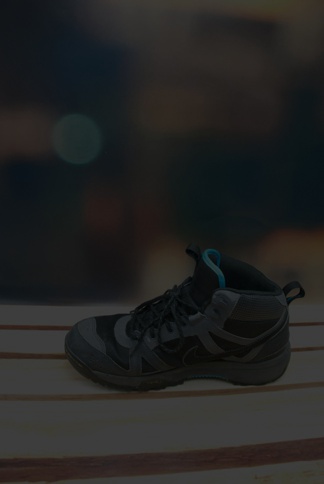}
    \rowname{Level 4}\populateRow{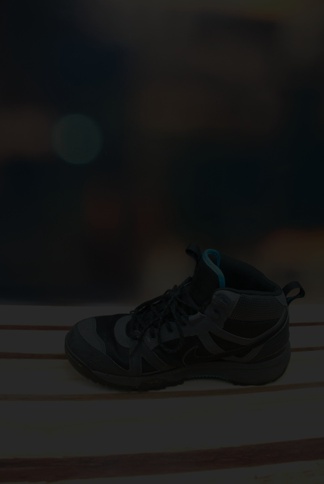}
    \rowname{Level 5}\populateRow{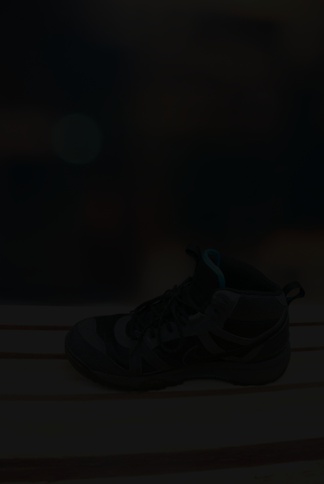}
\vspace{-4mm}
\caption{Sample images with challenging conditions (levels 1-5). The labels on the right side of each image are the top-1 output from Amazon Rekognition and Microsoft Azure Computer Vision. The labels in blue are considered correct for the selected object and those in yellow are considered incorrect.}
\label{fig:challenges_api_h}
\vspace{-4mm}
\end{figure*} 

\vspace{1mm}
\noindent{\bf Challenges types:}
 Images in the \texttt{CURE-OR} dataset were separated into color and grayscale images to investigate the importance of color information in object recognition. We processed original images to simulate realistic challenging conditions including \emph{underexposure}, \emph{overexposure}, \emph{blur}, \emph{contrast}, \emph{dirty lens}, \emph{salt and pepper noise}, and \emph{resizing}. Brightness of images was adjusted to obtain \emph{underexposure} and \emph{overexposure} images. Pixel values were remapped to obtain \emph{contrast} images and smoothed out with a Gaussian operator to obtain \emph{blur} images. \emph{Dirty lens} images were obtained by overlaying dirty lens patterns over original images. In the first \emph{dirty lens} category, a single dirt pattern was blended into images with weights that were proportional to challenge levels. In the second category, a distinct dirt pattern was overlaid for each challenge level. In the \emph{salt and pepper} category, pixels were randomly replaced with zeros and ones. \emph{Salt and pepper noise} was synthesized with scikit-image version 0.13.0 and all other challenging conditions were simulated with Python Imaging Library version 4.2.1.

\vspace{1mm}
\noindent{\bf Challenges levels:} The number of distinct challenge levels is 4 in \emph{resizing} category and 5 in all other categories. Challenge levels were adjusted based on visual inspection instead of numerical progression. Challenge level adjustment experiments were conducted by a group of three individuals, two experts and one intermediate in image processing, in an office space under standard lighting conditions. In the first experiment, we utilized a 65-inch LED TV display, whose settings were fixed to standard viewing conditions. We tuned the parameters in challenge generation such that challenges in level 1 were visible but they did not affect recognition, whereas in challenge level 5, objects were hardly recognizable or further changes in parameters did not correspond to significant perceivable difference. Mid challenge levels were linearly interpolated between min and max challenge levels. In the second experiment, we displayed level-1 and level-5 sample images on a 27-inch LCD PC monitor and observed that initial selection criteria still held in a different setup. In both experiments, subjects were located at a distance of twice the display size. Sample images corresponding to different challenge types and levels are shown in Fig. \ref{fig:challenges_api_h} with the API results that will be discussed in the following section.

\section{Recognition under Challenging Conditions}
\label{sec:exp}
There are numerous visual recognition applications that are publicly available, but it is not feasible to perform large-scale experiments in majority of these applications. After a preliminary feasibility study, we conducted our experiments using
\href{https://aws.amazon.com/rekognition/}{Amazon Rekognition API (AMZN)} and \href{https://azure.microsoft.com/en-us/services/cognitive-services/computer-vision/}{Microsoft Azure Computer Vision API (MSFT)}. Rekognition is a deep learning-based image recognition system provided by Amazon Web Services. Microsoft Azure Computer Vision utilizes image processing algorithms to analyze visual contents. We use the object recognition feature from AMZN and the image tagging functionality from MSFT to assess their performance.

\subsection{Experimental Setup}
\label{subsec:challenge_setup}


\noindent{\bf Test Set:} For both APIs, we input challenge-free images of 100 objects from CURE-OR dataset to collect object labels, and analyzed the collected labels semantically to identify the labels that can be considered ground truth. For example, a Coca Cola bottle can be recognized as a beverage, a bottle, a drink, a coke, a soda, or a glass. We selected 4 objects from each object category based on the number of images that can be correctly identified by each API. If a selected object was similar to another object in the same category, the object with more correctly identified images was utilized. Microsoft identified only 3 objects in health/personal care category, so one object with the least number of correctly identified images from Amazon was excluded for a fair analysis, which led to 23 objects (top objects) selected for each API. For direct comparison of the platforms, we also analyzed the recognition performance on the 10 common objects between the 23 top objects of both APIs. In total, we tested the performance of each API with $230,000$ images. 
We limited the number of tested objects to minimize the total cost of the experiment using Amazon. However, we were able to test the recognition performance for the remaining images using Microsoft thanks to promotional credits. On average, testing per image took 1.3 seconds for Amazon and 0.2 seconds for Microsoft.

\vspace{1mm}
\noindent{\bf Performance Metrics:} Top-5 accuracy metric was employed to measure the recognition performance. Outputs of both APIs were sorted based on their confidence levels and the highest 5 labels were selected for top-5 accuracy calculation. If multiple labels shared the same confidence level as the $5^{th}$ label, they were also included. If a ground truth label was among the top list, it was considered as a correct prediction. After repeating this procedure for all the images, we calculated the ratio between the number of correct predictions and the total number of object images to obtain the overall top-5 accuracy. In addition to the overall recognition performance in terms of top-5 accuracy, we also provided confusion matrices for different challenge levels to analyze misclassifications. Each row of the matrix corresponds to instances of actual categories and each column of the matrix corresponds to instances of predicted categories. An ideal classifier would have a confusion matrix with a highlighted main diagonal. Each element in the confusion matrix is colored based on the classification accuracy according to the legend next to each figure. Since both recognition APIs do not provide a comprehensive list of possible labels, the labels that did not belong to any categories were excluded.

\subsection{Recognition Performance}

\begin{figure}[htbp!]
\vspace{-2mm}
\begin{minipage}[b]{0.30\linewidth}
  \centering
\includegraphics[width=\textwidth]{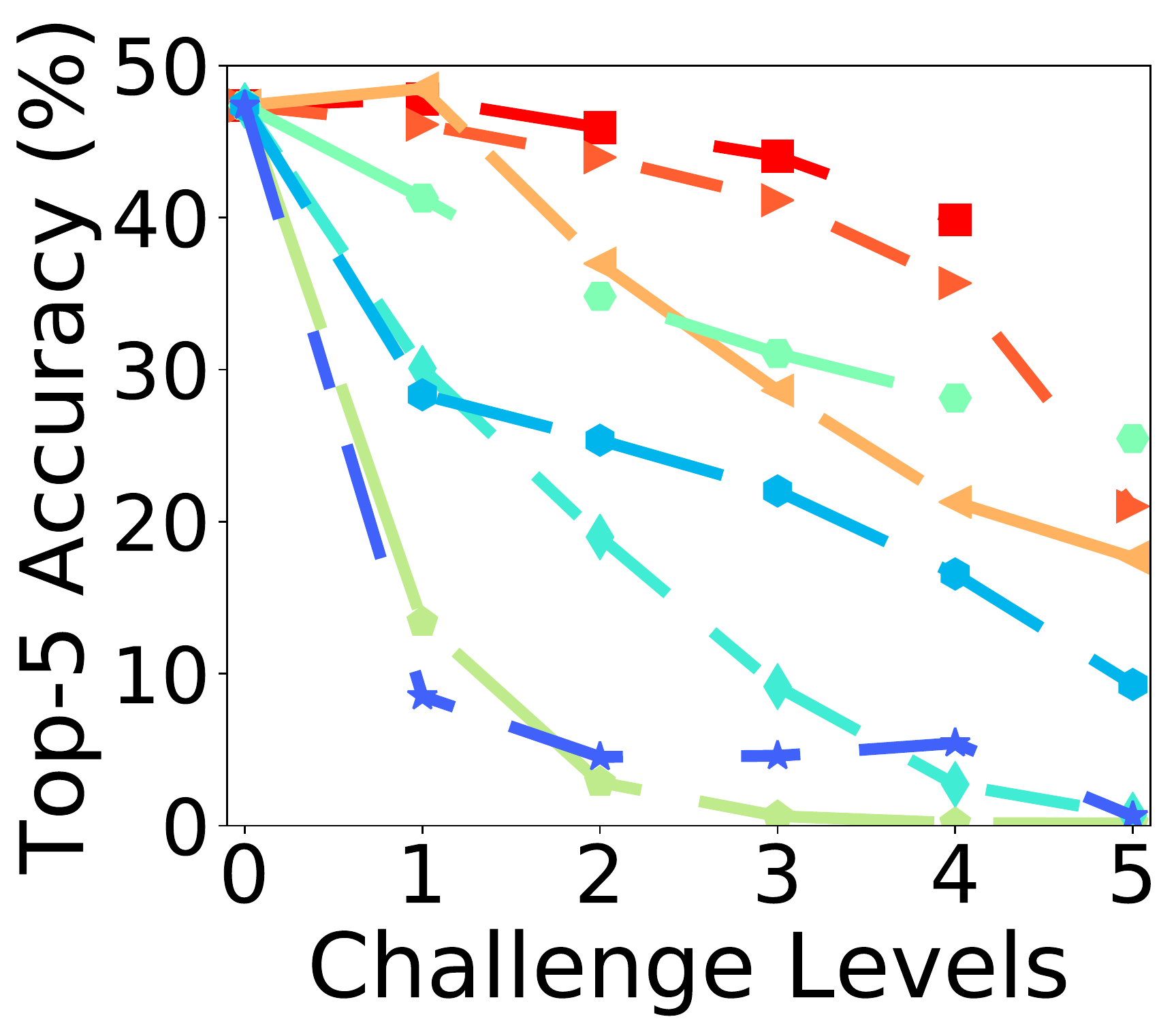}\vspace{-1mm}
  \centerline{\footnotesize{(a) AMZN: Top obj-C}}\vspace{1mm}
\end{minipage}
\hfill
\begin{minipage}[b]{0.30\linewidth}
  \centering
\includegraphics[width=\textwidth]{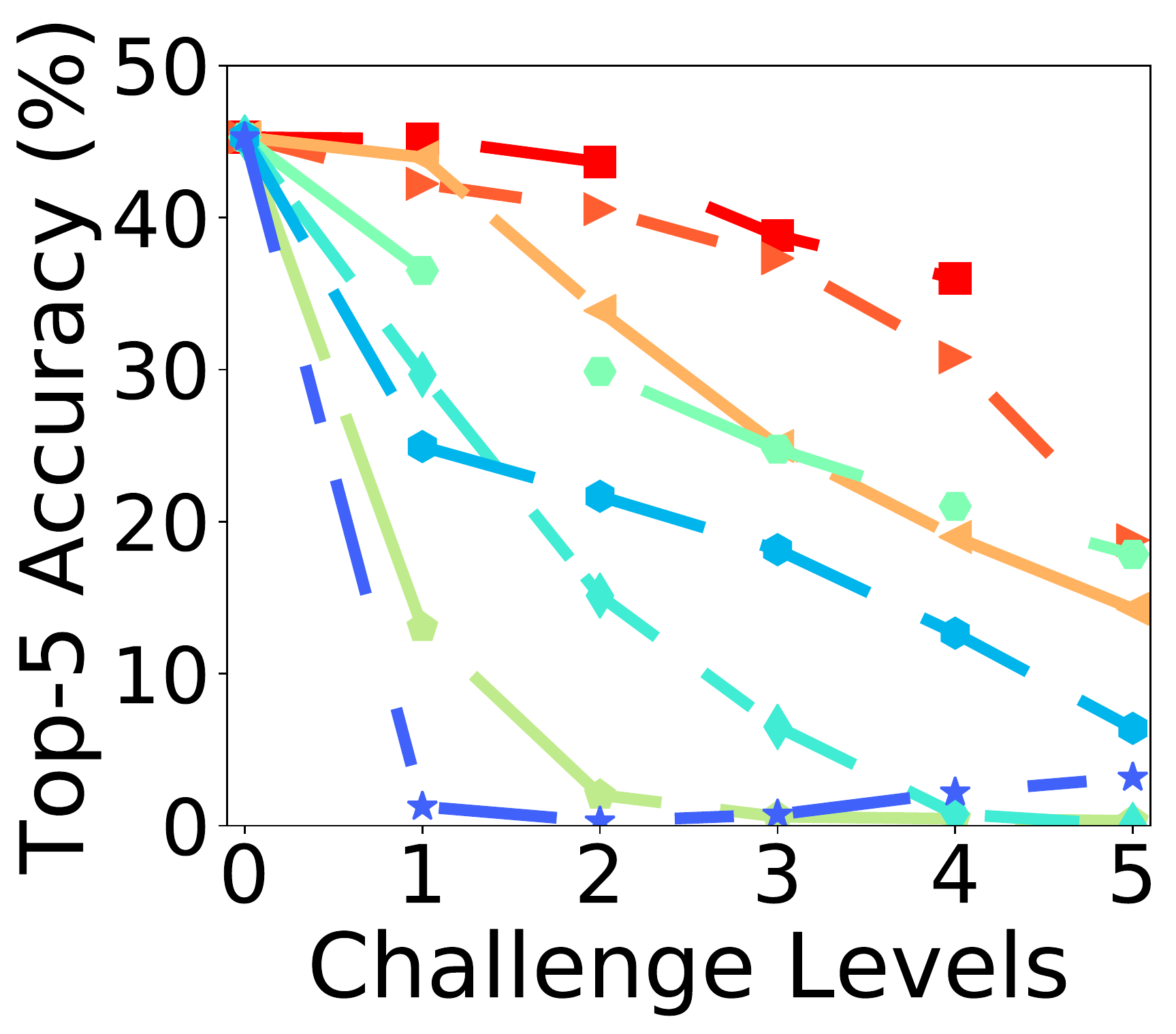}\vspace{-1mm}
  \centerline{\footnotesize{(b) AMZN: Top obj-G}}\vspace{1mm}
\end{minipage}
\hfill
\begin{minipage}[b]{0.30\linewidth}
  \centering
\includegraphics[width=\linewidth]{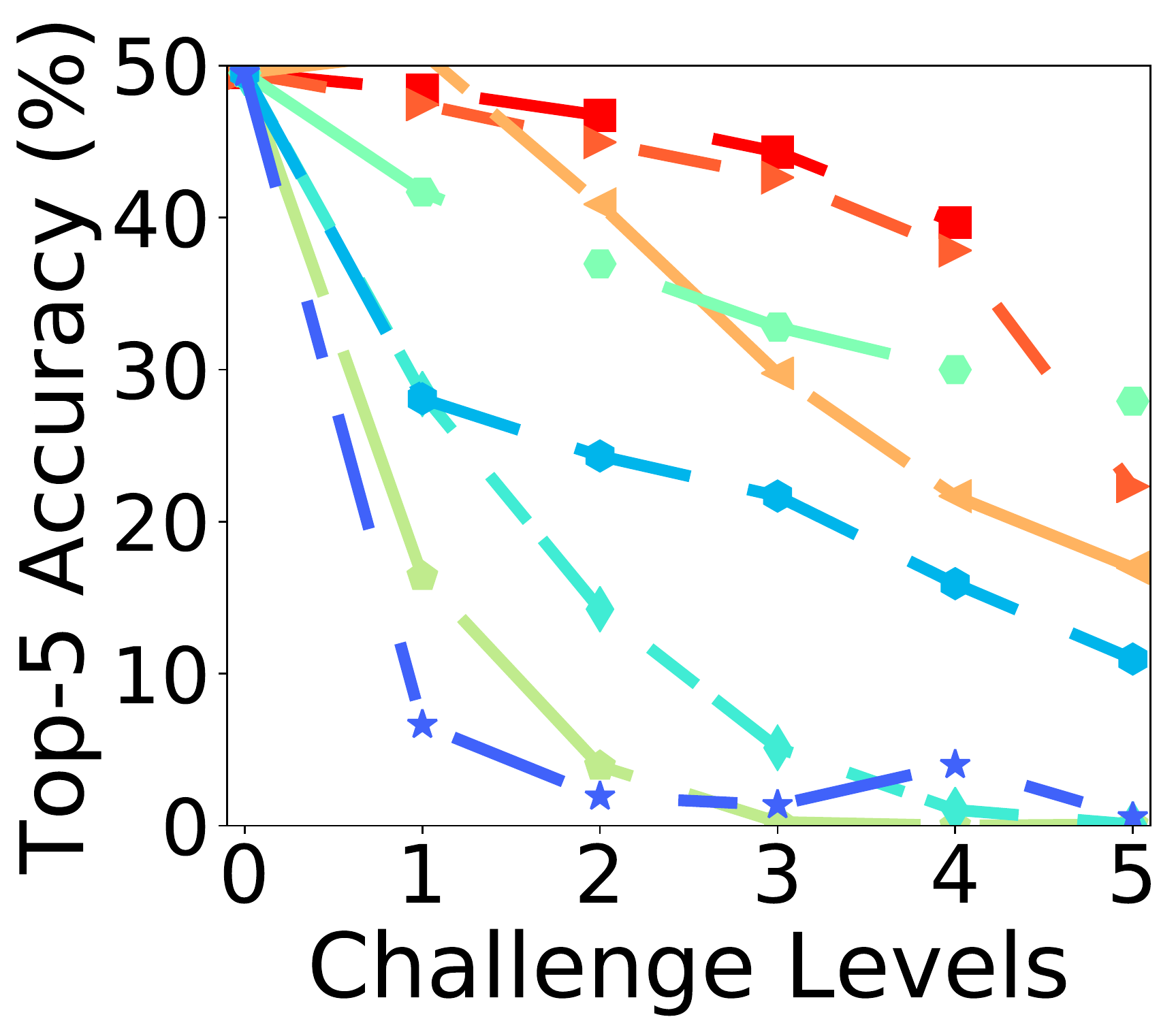}\vspace{-1mm}
  \centerline{\footnotesize{(c) AMZN: Common obj-C}}\vspace{1mm}
\end{minipage}
\begin{minipage}[b]{0.30\linewidth}
  \centering
\includegraphics[width=\linewidth]{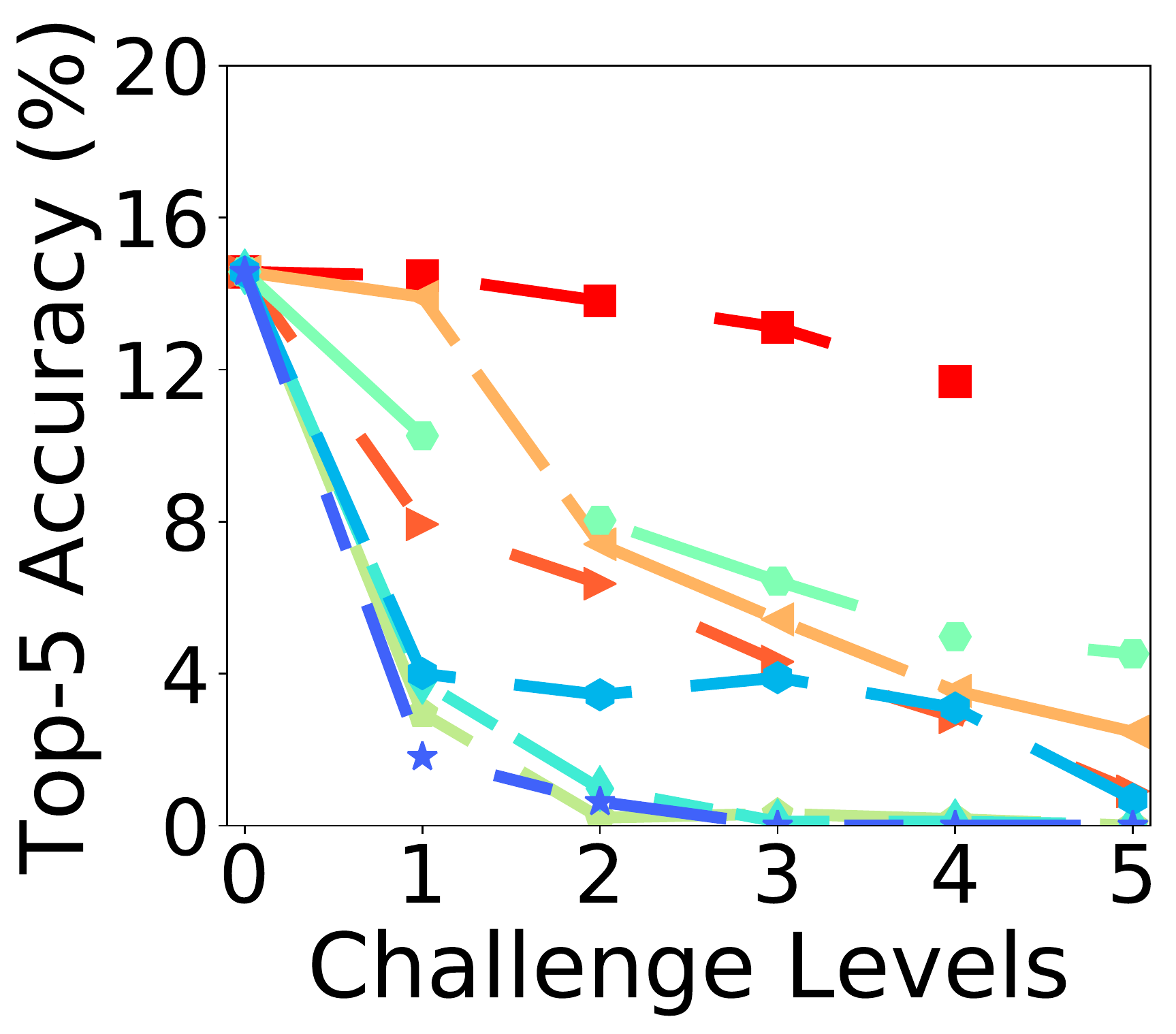}\vspace{-1mm}
  \centerline{\footnotesize{(d) MSFT: Top obj-C}}\vspace{2mm}
\end{minipage}
\hfill
\begin{minipage}[b]{0.30\linewidth}
  \centering
\includegraphics[width=\linewidth]{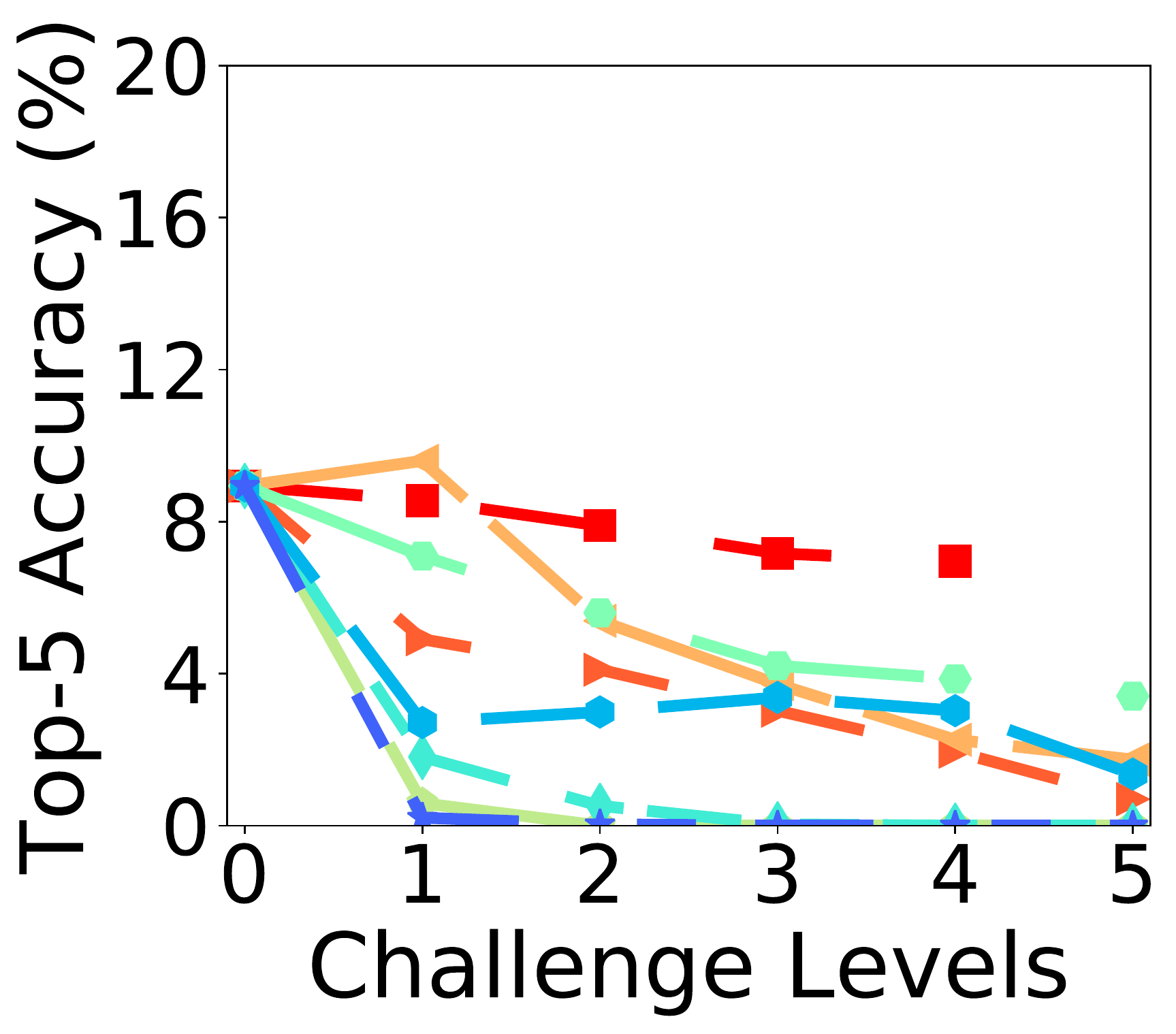}\vspace{-1mm}
  \centerline{\footnotesize{(e) MSFT: Top obj-G}}\vspace{2mm}
\end{minipage}
\hfill
\begin{minipage}[b]{0.30\linewidth}
  \centering
\includegraphics[width=\linewidth]{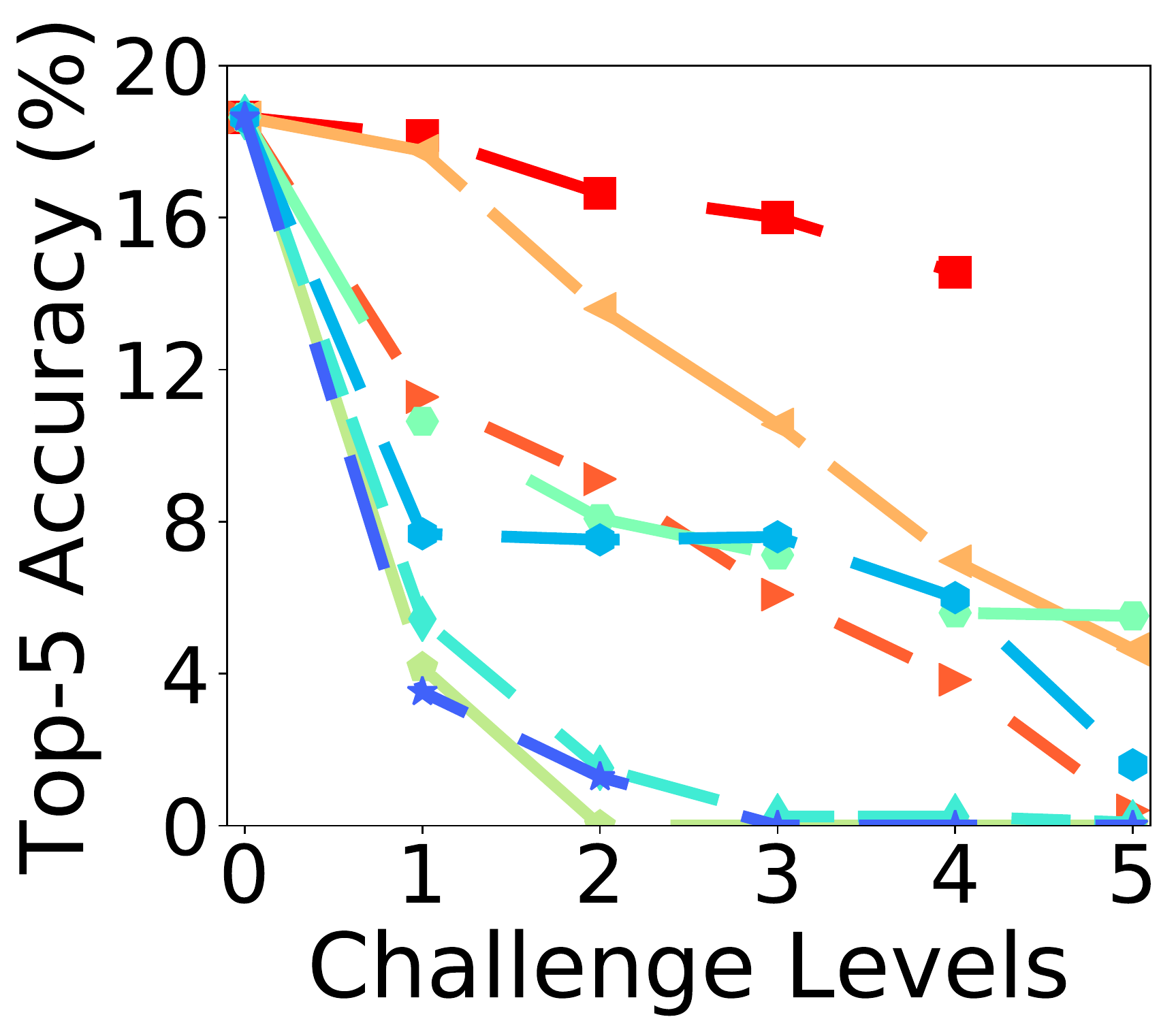}\vspace{-1mm}
  \centerline{\footnotesize{(f) MSFT: Common obj-C} }\vspace{2mm}
\end{minipage}
\centering
\includegraphics[width=\linewidth, trim= 10mm 5mm 10mm 0mm]{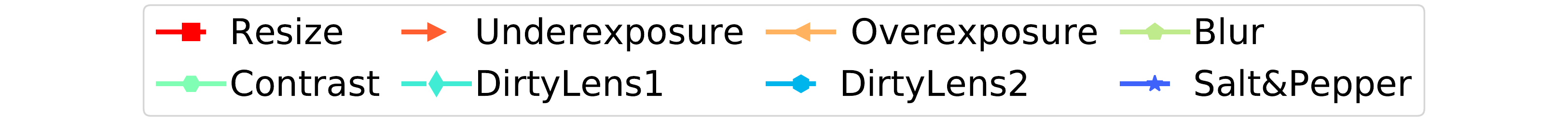}\vspace{1mm}
\caption{Accuracy of recognition platforms versus challenging conditions for color (C) and grayscale (G) images.}\vspace{-2mm}
\label{fig:challenges_accuracy}
\end{figure}
\begin{figure}[h]
\begin{minipage}[b]{0.23\linewidth}
  \centering
\includegraphics[width=\textwidth, trim = 5mm 5mm 5mm 2mm]{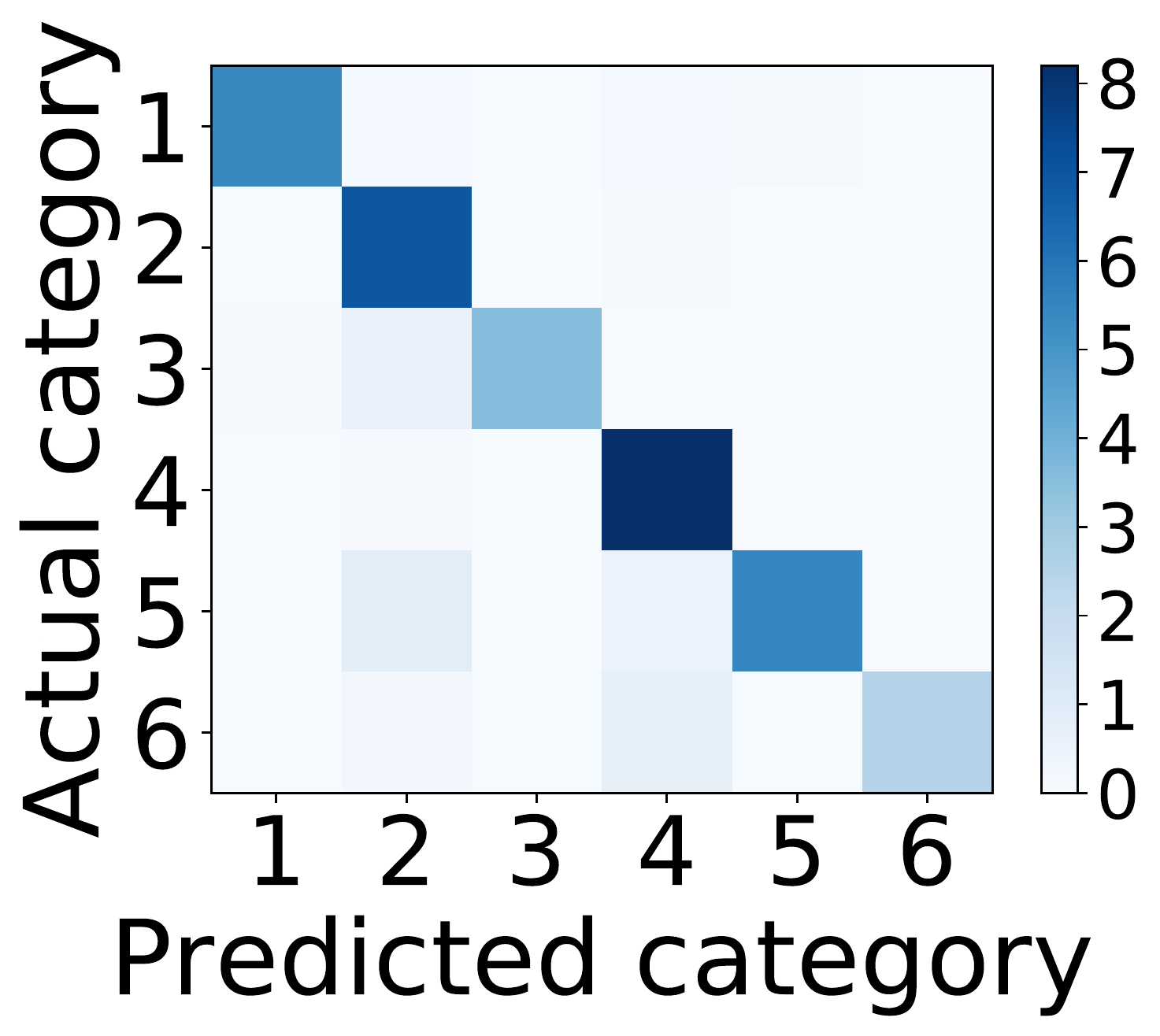}
  
  \centerline{\footnotesize{(a) AMZN: Lev\_0 }}\vspace{0mm}
\end{minipage}
\hfill
\begin{minipage}[b]{0.23\linewidth}
  \centering
\includegraphics[width=\textwidth, trim = 5mm 5mm 5mm 2mm]{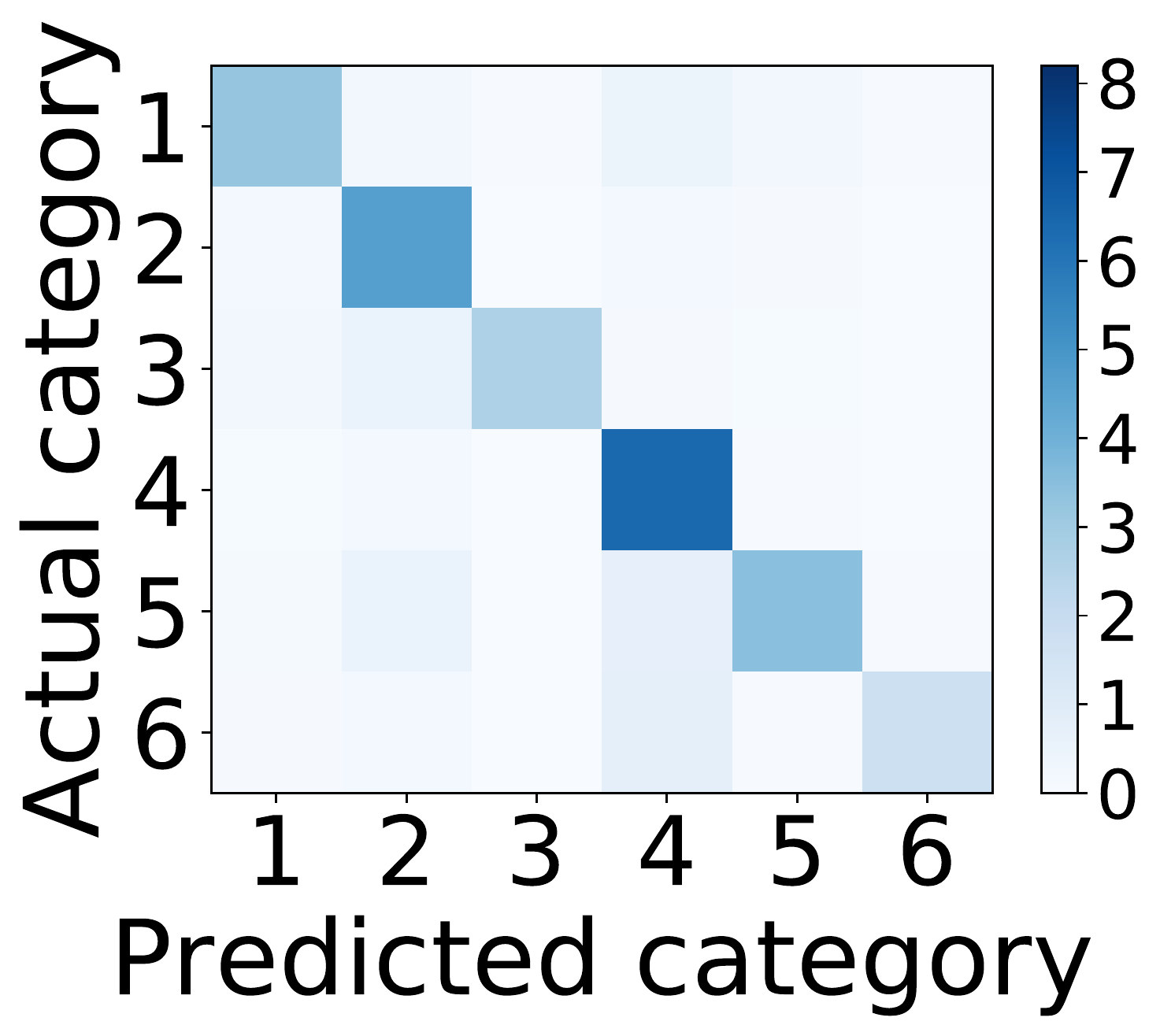}
  
  \centerline{\footnotesize{(b) AMZN: Lev\_1}}\vspace{0mm}
  \end{minipage}
\hfill
\begin{minipage}[b]{0.23\linewidth}
  \centering
  \includegraphics[width=\linewidth, trim = 5mm 5mm 5mm 2mm]{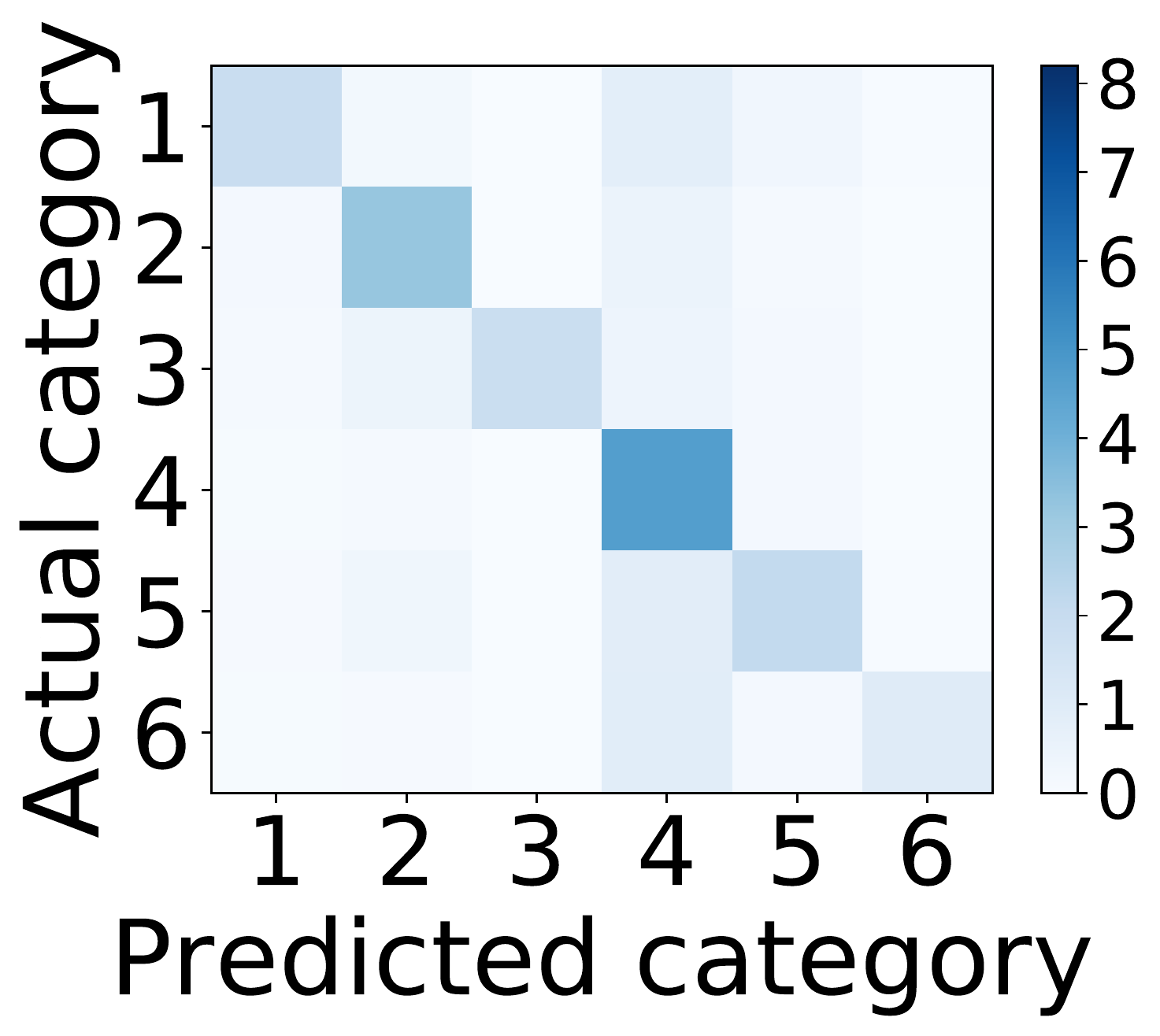}
  \vspace{0.01 cm}
  \centerline{\footnotesize{(c) AMZN: Lev\_3}}\vspace{0mm}
\end{minipage}
\hfill
\begin{minipage}[b]{0.23\linewidth}
  \centering
\includegraphics[width=\linewidth, trim = 5mm 5mm 5mm 2mm]{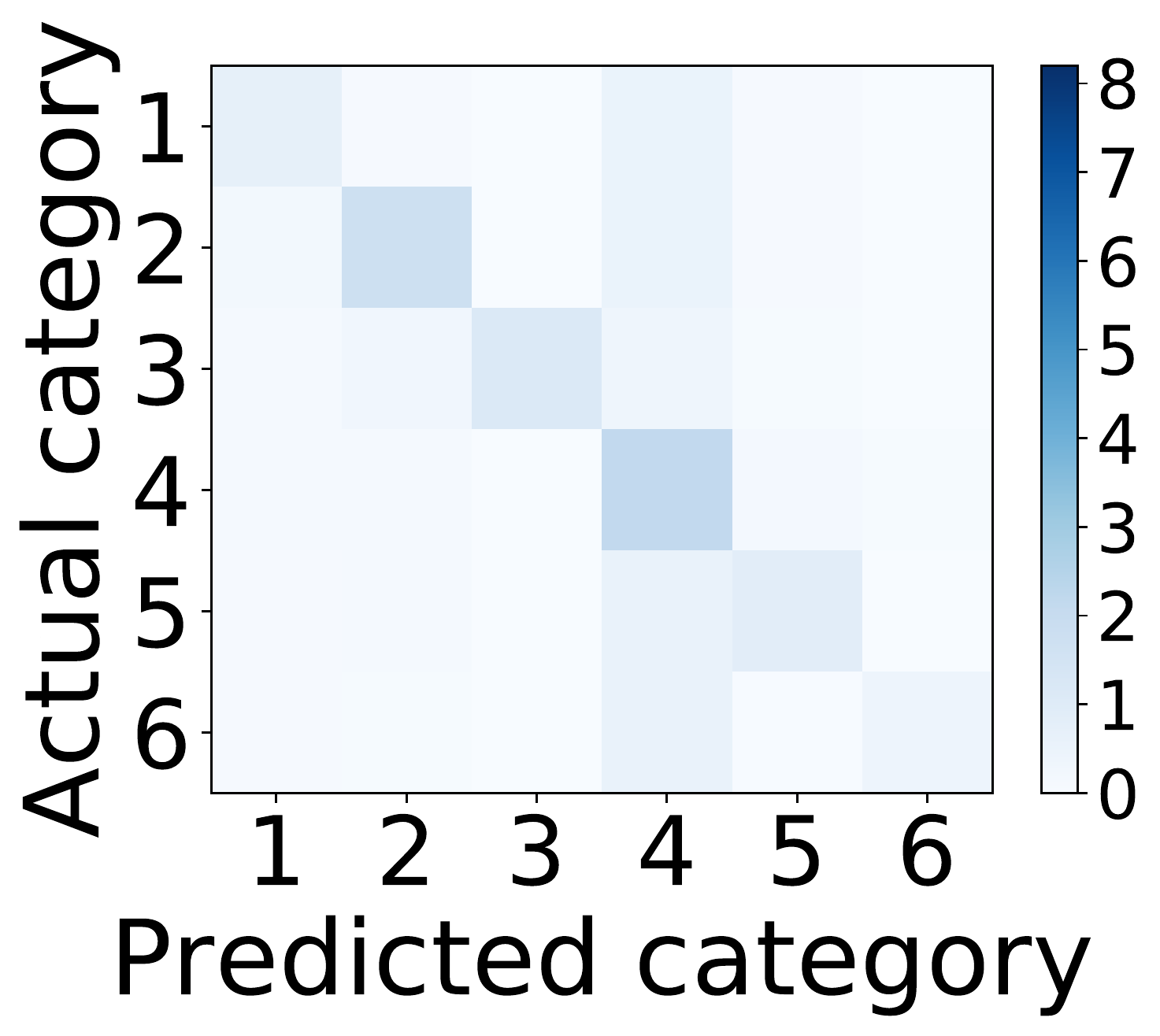}
  \vspace{0.01 cm}
  \centerline{\footnotesize{(d) AMZN: Lev\_5}}\vspace{0mm}
\end{minipage}
\hfill
\vspace{1mm}
\begin{minipage}[b]{0.23\linewidth}
  \centering
\includegraphics[width=\textwidth, trim = 5mm 5mm 5mm 5mm]{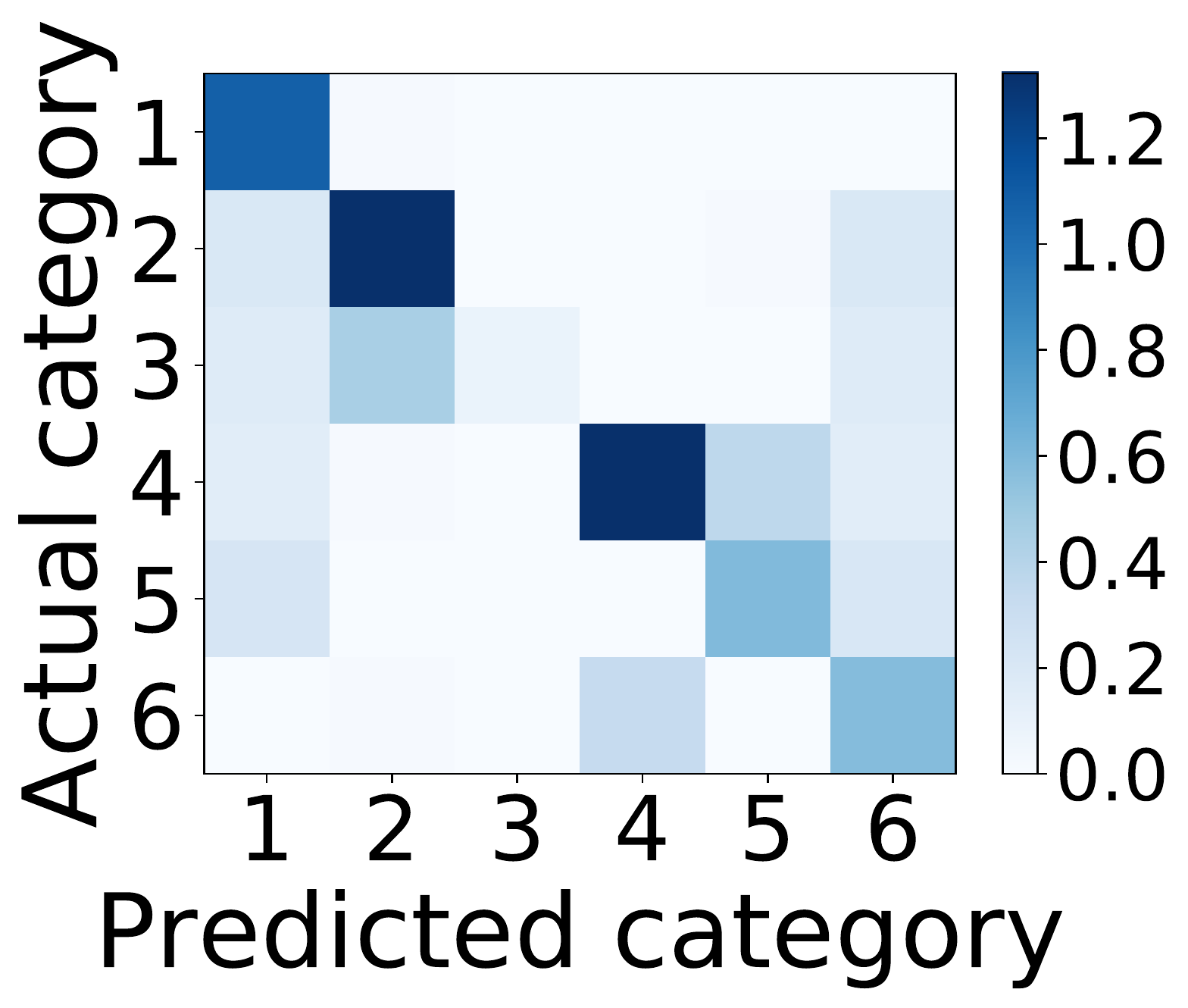}
  
  \centerline{\footnotesize{(e) MSFT: Lev\_0}}\vspace{0mm}
\end{minipage}
\hfill
\begin{minipage}[b]{0.23\linewidth}
  \centering
\includegraphics[width=\textwidth, trim = 5mm 5mm 5mm 5mm]{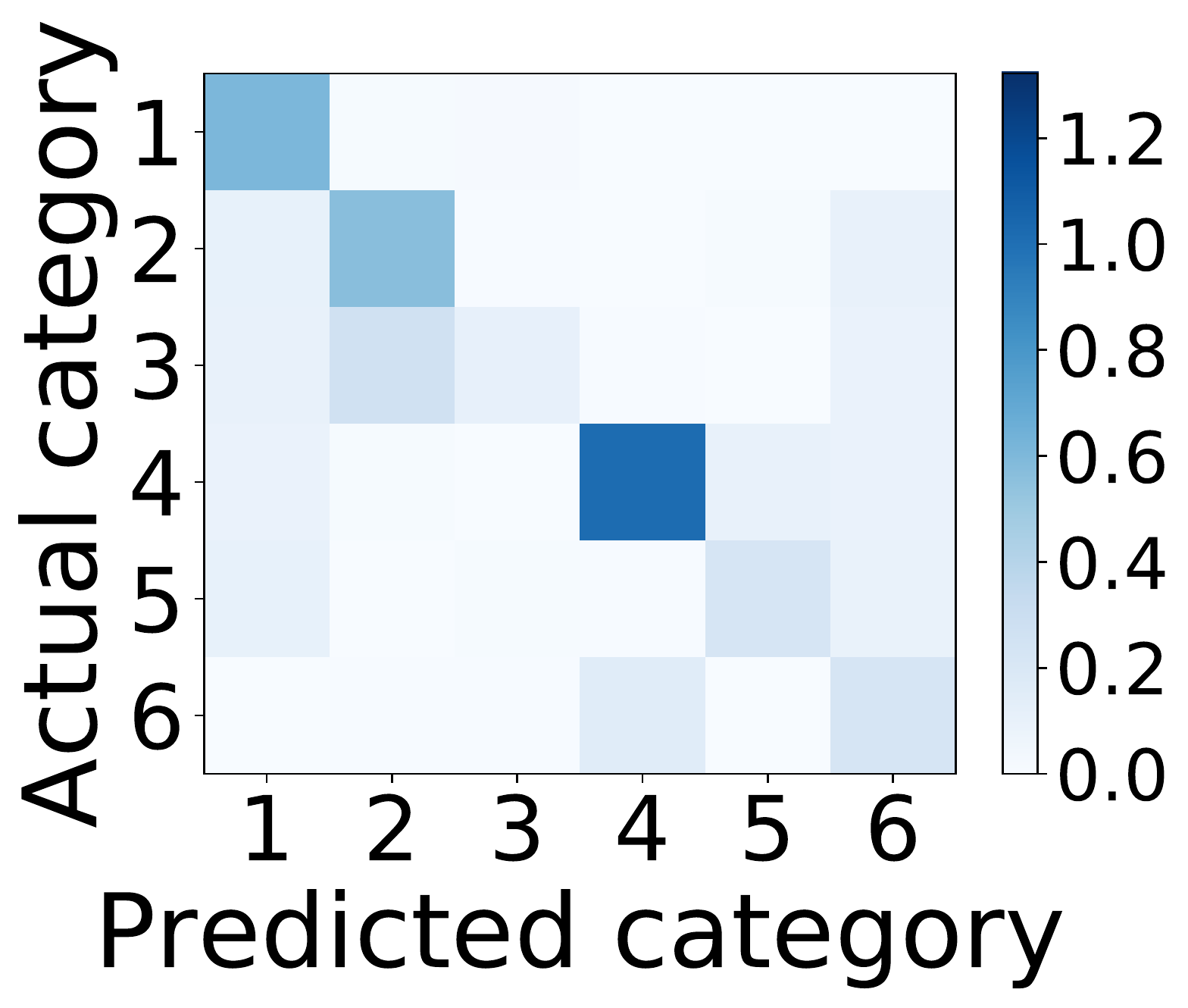}
  
  \centerline{\footnotesize{(f) MSFT: Lev\_1}}\vspace{0mm}
\end{minipage}
\hfill
\begin{minipage}[b]{0.23\linewidth}
  \centering
\includegraphics[width=\linewidth, trim = 5mm 5mm 5mm 5mm]{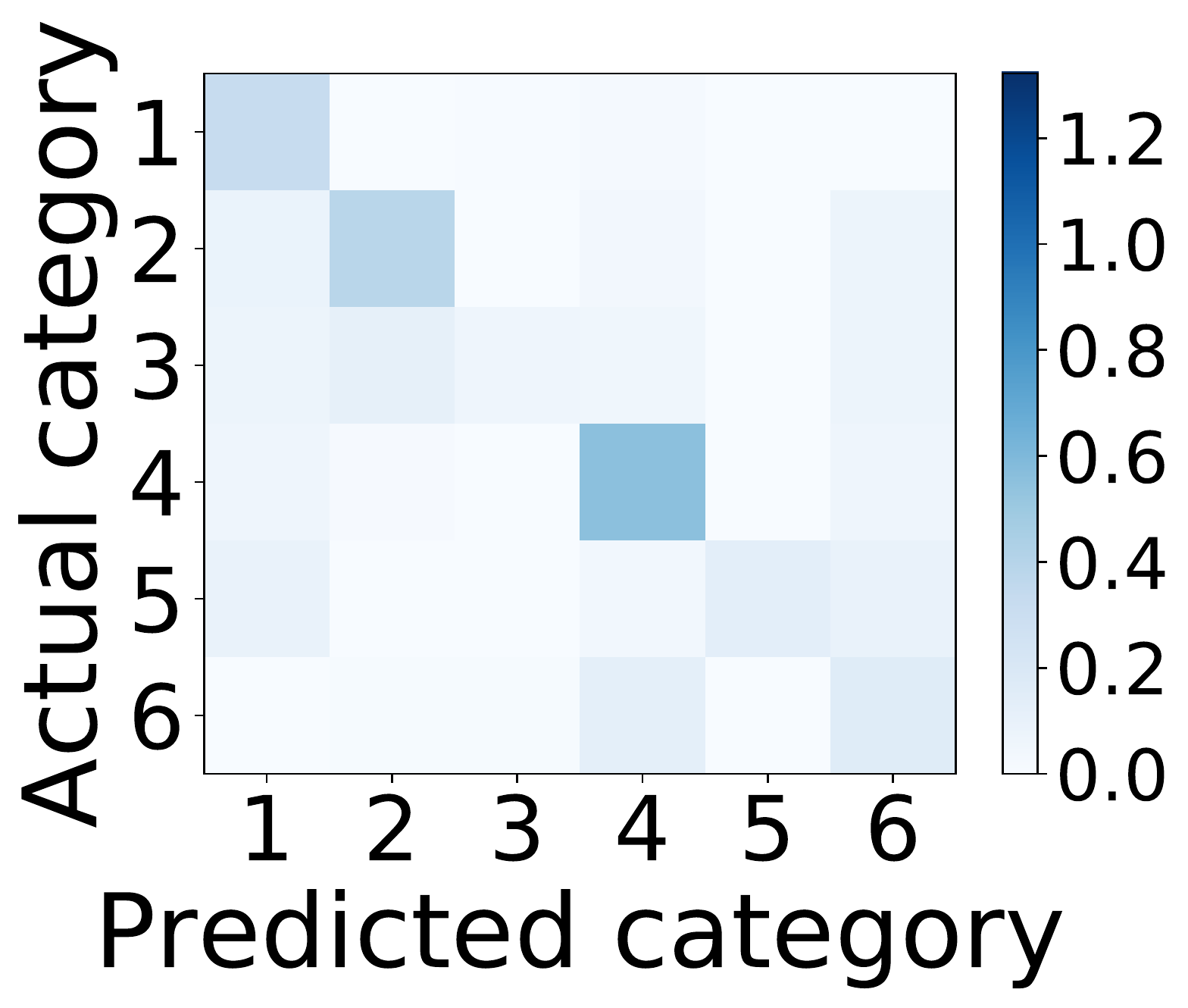}
  \vspace{0.01 cm}
  \centerline{\footnotesize{(g) MSFT: Lev\_3}}\vspace{0mm}
\end{minipage}
\hfill
\begin{minipage}[b]{0.23\linewidth}
  \centering
\includegraphics[width=\linewidth, trim = 5mm 5mm 5mm 5mm]{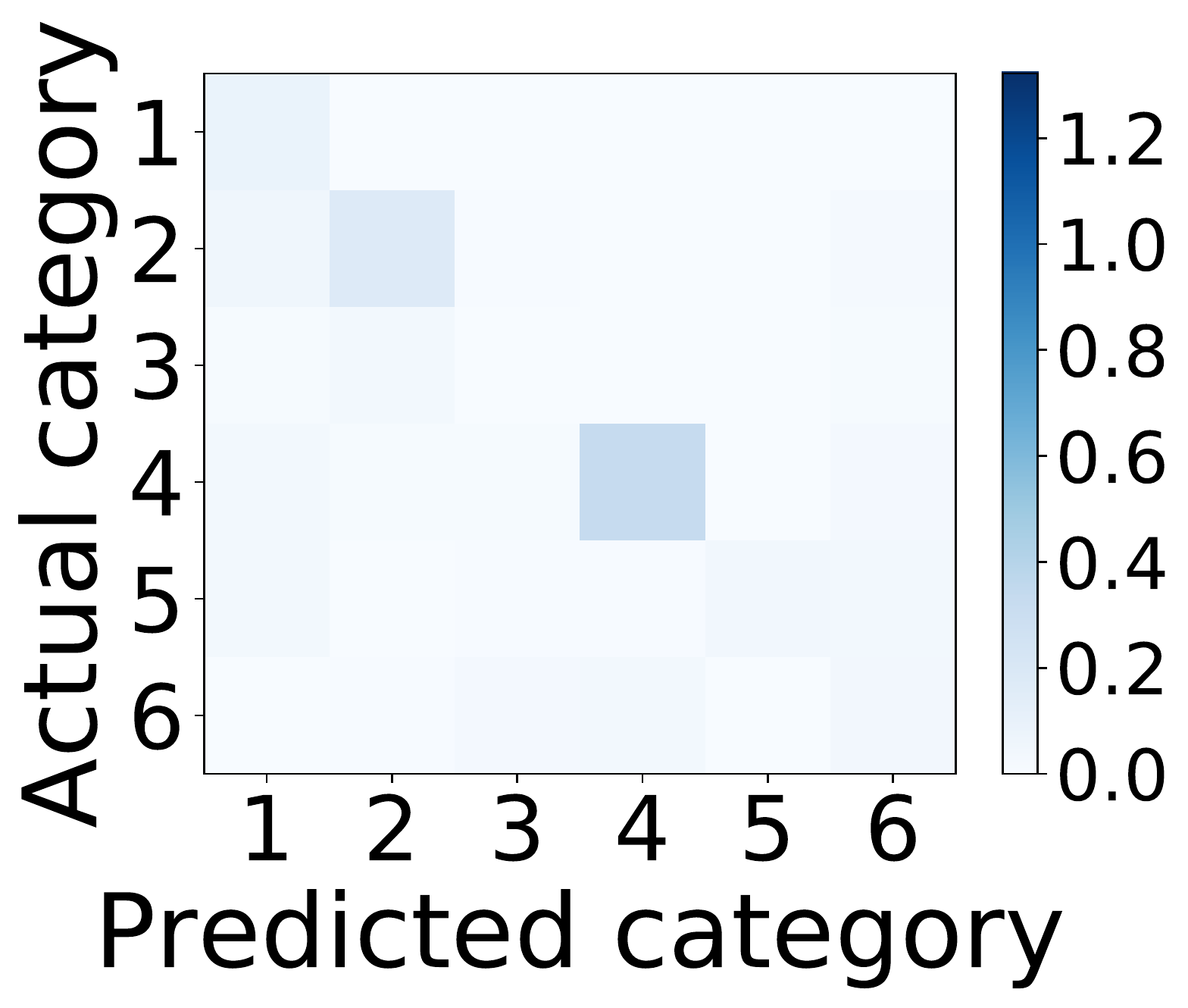}
  \vspace{0.01 cm}
  \centerline{\footnotesize{(h) MSFT: Lev\_5}}\vspace{0mm}
  \end{minipage}
\centering
\vspace{-1mm}\includegraphics[width=\linewidth, clip, trim=10mm 2mm 25mm 0mm ]{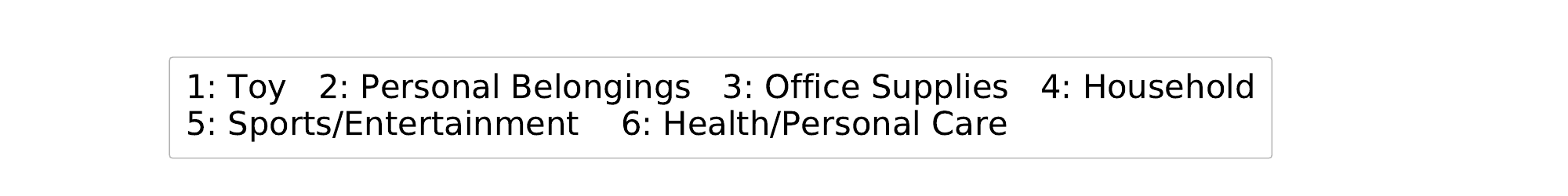}
\caption{Confusion matrix for object category classification at different challenge levels.}\vspace{-5mm}
\label{fig:confusion_mat_common}
\end{figure}

In this section, we analyze the performance of both recognition platforms under challenging conditions. We report the recognition performance using color and grayscale images of top and common objects in Fig. \ref{fig:challenges_accuracy}. Moreover, we discuss the misclassification between object categories under varying challenge levels with confusion matrices in Fig. \ref{fig:confusion_mat_common}. Lastly, we report the overall recognition performance of both platforms in Table \ref{tab_overall_results}.

\noindent{\bf Color versus Grayscale:} To understand the contribution of color in terms of recognition performance, Fig. \ref{fig:challenges_accuracy}(a) can be compared with Fig. \ref{fig:challenges_accuracy}(b) for Amazon, and Fig. \ref{fig:challenges_accuracy}(d) with Fig. \ref{fig:challenges_accuracy}(e) for Microsoft. Based on the results in Fig. \ref{fig:challenges_accuracy}, we observe that both recognition platforms utilize color information. However, color information affects overall recognition performance more significantly for Microsoft compared to Amazon.


\vspace{1mm}
\noindent{\bf Challenging Conditions:} For both platforms, \emph{resizing} affects the performance the least for the 4 challenge levels present. \emph{Blur}, \emph{salt \& pepper} noise, and \emph{dirty lens 1} challenging conditions are the most effective in degrading the performance, leading to less than 1\% top-5 accuracy for Amazon and 0.1\% for Microsoft. In general, the challenges that remove structural components in an image degrade the application performance significantly. \emph{Blur} filters out high-frequency components and \emph{salt \& pepper} noise replaces random pixel values that corrupt structural components, which are critical for recognition task. Even though \emph{resizing} also filters out high-frequency components, the level of smoothing is not as remarkable as \emph{blur} and most structural components are preserved. \emph{Dirty lens 1} challenge overlays a dirty lens pattern with different weights corresponding to challenge levels, which effectively block edge and texture in an image. We also report the recognition performance using color images of 10 common objects in Fig.~\ref{fig:challenges_accuracy}(c) and Fig.~\ref{fig:challenges_accuracy}(f). Similar to top object performance, Amazon significantly outperforms Microsoft in all categories.

\vspace{1mm}
\noindent{\bf Misclassification:} We analyze the object categorical misclassifications at different challenge levels with confusion matrices as shown in Fig. \ref{fig:confusion_mat_common}, where the first row corresponds to Amazon Rekognition and the second row corresponds to Microsoft Azure. We observe that the main diagonals of Amazon Rekognition are more predominant than the diagonals of Microsoft Azure at every challenge level. As the challenge level increases, highlights in the main diagonals significantly diminish, which shows the vulnerabilities of the recognition platforms under challenging conditions. 

\vspace{1mm}
\noindent{\bf Overall Performance:} We report the overall recognition performance of Amazon Rekognition and Microsoft Azure in Table \ref{tab_overall_results}. Test sets include challenge images as well as original images and performance is reported in terms of top-5 accuracy. For Amazon Rekognition, we report the results for 10 common objects and 23 top objects as described in Section~\ref{subsec:challenge_setup}. For Microsoft Azure, we also report the results for the entire \texttt{CURE-OR} dataset. Amazon Rekognition shows significantly higher performance compared to Microsoft Azure for the top and common objects in both color and grayscale. Both recognition platforms show slightly better accuracy for the 10 common objects than the 23 top objects because the objects that can be recognized by both platform lead to a slightly higher average accuracy compared to remaining objects from top objects. When all objects are considered, Microsoft Azure's performance degrades even more, getting closer to 1\% top-5 accuracy because majority of the remaining 77 objects are not accurately recognized even when they are challenge-free. 

\begin{table}[b]\vspace{-3mm}
\footnotesize 
\centering
\caption{Overall recognition performances.}
\label{tab_overall_results}
\begin{tabular}{c|c|c|c|c|c|c}
\hline
\multirow{2}{*}{} & \multicolumn{3}{c|}{Color} & \multicolumn{3}{c}{Grayscale} \\ \cline{2-7} 
 & Top & Common & All & Top& Common& All \\ \hline
\begin{tabular}[c]{@{}c@{}}AMZN\end{tabular} & 25.78 & 26.03 & - & 22.78 & 23.6 & - \\ \hline
\begin{tabular}[c]{@{}c@{}}MSFT\end{tabular} & 5.66 & 7.62 & 1.83 & 3.58 & 5.17 & 1.01 \\ \hline
\end{tabular}
\end{table}

\vspace{1mm}
\section{Recognition Accuracy versus Image Quality}
\label{sec:quality}
\subsection{Experimental Setup}

\noindent{\bf Image Quality Algorithms, Databases and Distortions:} We utilize three different types of full-reference image quality assessment algorithms including PSNR as a fidelity measure, SSIM \cite{Wang2004} as a structural similarity measure, and UNIQUE \cite{Temel_UNIQUE} as a data-driven quality measure based on color and structure. We validate the subjective quality estimation performance over LIVE \cite{live}, MULTI \cite{multi}, and TID13 \cite{tid} databases, which include $4,432$ images and $30$ distortion types. Distortion types in test databases can be grouped into 7 main categories as \emph{lossy compression}, \emph{image noise}, \emph{communication error}, \emph{image blur}, \emph{color artifacts}, \emph{global distortions} (intensity shift/contrast change), and \emph{local distortions}.

\vspace{1mm}
\noindent{\bf Estimation Performance Calculation:} In this study, we estimate the recognition accuracy with respect to changes in photometric conditions. The top-5 accuracy results from Amazon Rekognition and Microsoft Azure were averaged for this section. At first, we identify the challenge-free images as references, which lead to the highest recognition accuracy. Then, we investigate the relationship between image quality and recognition accuracy. Specifically, first, we group images that are captured with a particular  device (5) in front of a specific backdrop (3) under a specific challenge level (6), which leads to 90 image groups for each challenge type. Each image group includes images of 10 common objects described in Section \ref{subsec:challenge_setup} for all orientations. Second, we calculate average recognition performance and image quality scores for each image group. Overall, for each challenge category, we obtain 90 image quality and recognition accuracy scores, which correspond to the average of image groups. Real-world living room and office backgrounds are excluded because performance in these backgrounds are significantly different from other backgrounds, which cannot be captured with quality estimators that require pixel-to-pixel correspondence. Third, we calculate the Spearman correlation between objective quality and recognition accuracy scores to obtain recognition accuracy estimation performance. Moreover, we calculate the Spearman correlation between objective and subjective scores in image quality databases (LIVE, MULTI, and TID13) to obtain subjective quality assessment performance of image quality assessment (IQA) algorithms.

\begin{table}[htbp!]
\small
\centering
\caption{Spearman correlation between average recognition accuracies and objective quality scores in the \texttt{CURE-OR} dataset.}
\label{tab_perf_recognition}
\begin{tabular}{c|ccc} \hline \hline
\textbf{Challenge Type} & \textbf{PSNR}& \textbf{SSIM}  & \textbf{UNIQUE}   \\ \hline \hline 
\bf Resize   & \bf0.755 & 0.382 & 0.312    \\
\bf Underexposure    & 0.722 & \bf0.807 & 0.407   \\ 
\bf Overexposure   & \bf0.642 & 0.432 & 0.576 \\ 
\bf Blur   & 0.782 & 0.749 & \bf0.876 \\ 
\bf Contrast   & \bf0.813 & 0.644 & 0.767 \\ 
\bf Dirty lens 1  & \bf0.915 & 0.872 & 0.841  \\ 
\bf Dirty lens 2  & \bf0.911 & 0.667 & 0.643  \\ 
\bf Salt \& Pepper   & \bf0.792 & 0.791 & 0.780 \\ \hline
\end{tabular}
\end{table}
 \vspace{-2mm}

\subsection{Recognition Accuracy and Subjective Quality Estimation}
\begin{figure}[htbp!]
\centering
    \includegraphics[clip, trim=2mm 2mm 1.5mm 1mm, width=0.32\linewidth]{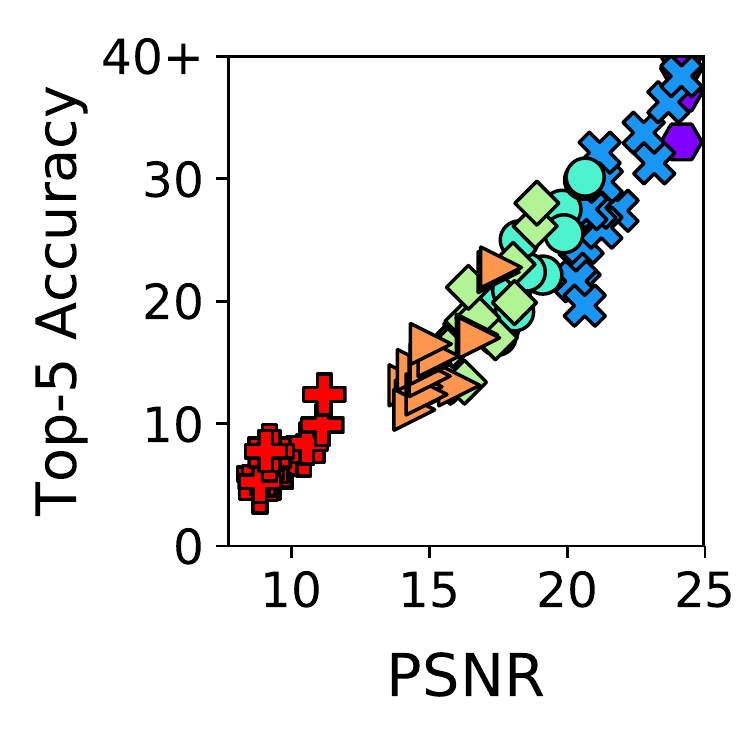}
    \includegraphics[clip, trim=2mm 2mm 1.5mm 1mm, width=0.32\linewidth]{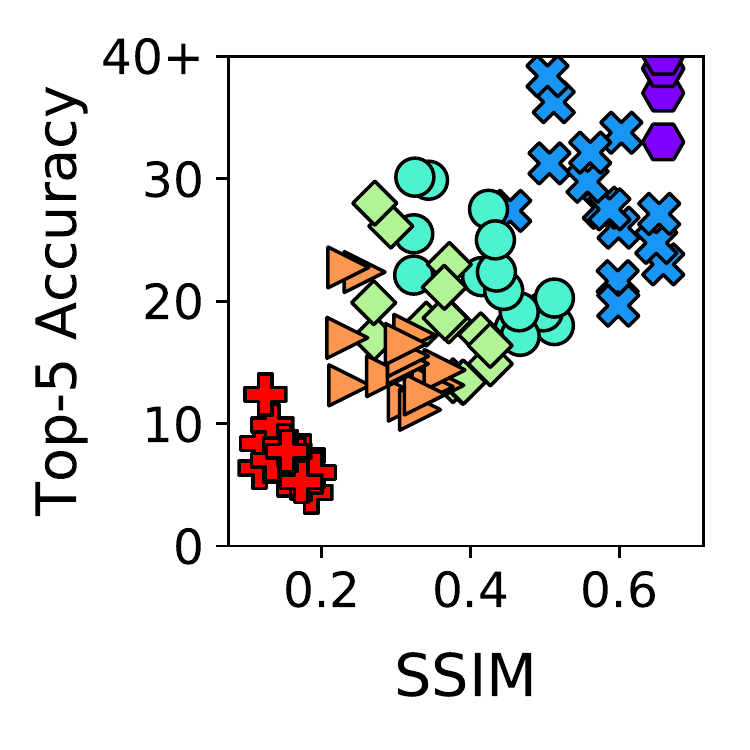}
    \includegraphics[clip, trim=2mm 2mm 1.5mm 1mm, width=0.32\linewidth]{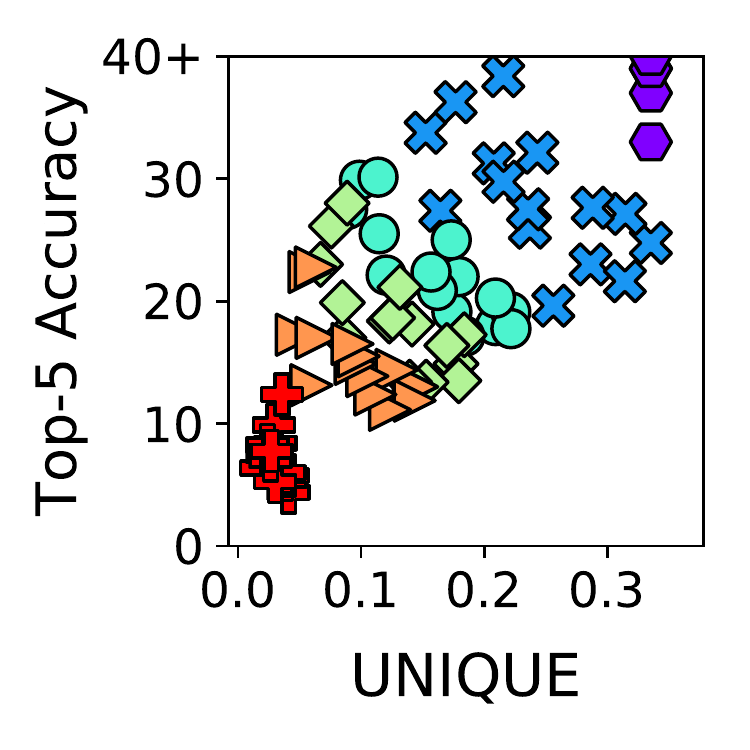}
    \includegraphics[clip, trim=1.3cm 5mm 1.1cm 0mm, width=0.98\linewidth]{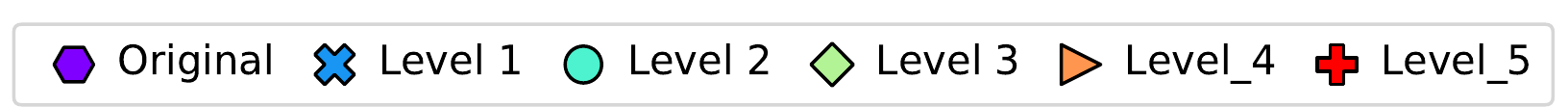}    
    \caption{Scatter plots of object recognition accuracy versus objective quality scores.}
    \label{fig:scatter}
    \vspace{-3mm}
\end{figure}

We report the recognition accuracy estimation in Table~\ref{tab_perf_recognition}  and subjective quality estimation in Table~\ref{tab_perf_quality}. In terms of challenge categories, PSNR is leading in majority of the categories whereas SSIM is leading in \emph{underexposure} category and UNIQUE in \emph{blur} category. Overall,  fidelity-based PSNR has the highest correlation, SSIM is the second measure, and UNIQUE is the third measure. Moreover, the relationship between the average recognition accuracy of both APIs on common objects and the quality scores can be observed from the scatter plots in Fig.~\ref{fig:scatter}. Scatter point distribution has a linear characteristic in PSNR whereas scatter points are more spread out for other quality measures. Even though SSIM and UNIQUE do not show a linear behavior, we can observe that scatter points corresponding to different challenge levels form distinct clusters. Therefore, UNIQUE and SSIM scores combined with top-5 accuracy scores can be informative in terms of identifying challenge level of the scenes.

According to subjective quality estimation performances in the image quality databases reported in Table~\ref{tab_perf_quality},  UNIQUE leads to highest correlation in majority of the distortion types and SSIM leads in \emph{compression} and \emph{global} categories. SSIM and UNIQUE are specifically designed for estimating subjective quality of images whereas PSNR is used for calculating the fidelity of a general signal representation based on the ratio of maximum signal power and noise power. Therefore, UNIQUE and SSIM outperform PSNR in the application they are designed for whereas they underperform in estimating recognition accuracy. Even though objective quality assessment algorithms can estimate the recognition performance in certain scenarios, they are inherently limited because significant perceptual changes captured by quality assessment algorithms may not be critical for recognition algorithms.

\begin{table}[htbp!]
\small
\centering
\caption{Spearman correlation between subjective quality scores and objective quality scores in the IQA databases.}
\label{tab_perf_quality}
\begin{tabular}{c|ccc} \hline \hline
\textbf{Challenge Type} & \textbf{PSNR}& \textbf{SSIM}  & \textbf{UNIQUE}   \\ \hline \hline 
\bf Compression   &0.871  &\bf 0.938    &0.934  \\ 
\bf Noise   &0.774  & 0.862  &\bf 0.875  \\ 
\bf Communication   &0.863  &0.931   &\bf 0.933  \\ 
\bf Blur   &0.803  & 0.905 &\bf 0.909  \\ 
\bf Color   &0.815  &0.235    & \bf 0.909 \\ 
\bf Global   &0.340  &\bf 0.499    & 0.356 \\ 
\bf Local   &0.543  &0.288   &\bf  0.645 \\ \hline
\end{tabular}
\vspace{-3mm}
\end{table}

\section{Conclusion}
\label{sec:conc}
In this paper, we have introduced a large-scale object recognition dataset denoted as \texttt{CURE-OR} whose images were captured with multiple devices in different backgrounds and orientations. We have utilized the introduced dataset to test the robustness of recognition APIs---Amazon Rekognition and Microsoft Azure Computer Vision---under challenging conditions. Based on the experiments, we have shown that both APIs are significantly vulnerable to challenging conditions that corrupt or eliminate high frequency components of an image and less vulnerable to conditions that affect pixel range. We have further analyzed the categorical misclassifications and concluded that recognition performance degrades for each challenge category as the challenge level increases. Overall, tested APIs resulted in a relatively low performance, which indicates the need for more accurate and robust solutions that are less sensitive to challenging conditions. Moreover, we also investigated the relationship between object recognition and image quality and showed that objective quality algorithms can estimate recognition performance under certain photometric challenging conditions if environmental conditions do not change significantly in terms of orientation and background.

\end{document}